\pdfoutput=1

\documentclass[11pt]{article}

\usepackage[]{ACL2023}

\usepackage{times}
\usepackage{latexsym}
\usepackage{amsmath}
\usepackage{amsfonts}
\usepackage{graphicx}
\usepackage{amssymb}
\usepackage{pifont}
\usepackage{hyperref}
\usepackage[nameinlink]{cleveref}
\usepackage{soul}
\usepackage{algorithm}
\usepackage{algpseudocode}
\usepackage{booktabs}
\usepackage{multirow}
\usepackage{arydshln}
\usepackage{subfigure}
\usepackage{lipsum}
\usepackage{xcolor}
\usepackage{CJKutf8}  

\newcommand{\framework}{\textsc{ConFit}}
\newcommand{\second}[1]{\textcolor{gray}{\emph{#1}}}
\newcommand{\bestunderline}[1]{{\textbf{#1}}}
\newcommand{\chinese}[1]{\begin{CJK}{UTF8}{gbsn}#1\end{CJK}}

\usepackage[T1]{fontenc}

\usepackage[utf8]{inputenc}

\usepackage{microtype}

\usepackage{inconsolata}

\title{\framework{}: Improving Resume-Job Matching using Data Augmentation and Contrastive Learning}

\author{Xiao Yu$^\dagger$~~~ 
        Jinzhong Zhang$^{\ddagger}$\label{author:intellipro}~~~
        Zhou Yu$^\dagger$~~~\\[3pt]
  $^\dagger$Columbia University~~~ 
  $^\ddagger$Intellipro Group Inc.\\[3pt] 
  \texttt{\{xy2437,zy2416\}@columbia.edu}\\
  \texttt{\{jinzhong\}@intelliprogroup.com}
}

\begin{document}
\maketitle
\begin{abstract}
A reliable resume-job matching system helps a company find suitable candidates from a pool of resumes, and helps a job seeker find relevant jobs from a list of job posts.
However, since job seekers apply only to a few jobs, interaction records in resume-job datasets are sparse.
Different from many prior work that uses complex modeling techniques, we tackle this sparcity problem using data augmentations and a simple contrastive learning approach.
\framework{} first creates an augmented resume-job dataset by paraphrasing specific sections in a resume or a job post.
Then, \framework{} uses contrastive learning to further increase training samples from $B$ pairs per batch to $\mathcal{O}(B^2)$ per batch.
We evaluate \framework{} on two real-world datasets and find it outperforms prior methods (including BM25 and OpenAI text-ada-002) by up to 19\% and 31\% absolute in nDCG@10 for ranking jobs and ranking resumes, respectively.\footnote{
We will release our code upon acceptance.
}
\end{abstract}

\section{Introduction}
Online recruitment platforms, such as LinkedIn, have over 900 million users, with over 100 million job applications made each month \cite{linkedin}. With the ever increasing growth of online recruitment platforms, building \emph{fast} and \emph{reliable} person-job fit systems is desiderated. A practical system should be able to quickly select suitable talents and jobs from large candidate pools, and also reliably quantify the ``matching degree'' between a resume and a job post.

Since both resumes and job posts are often stored as text data, many recent work \cite{pjfnn,APJFNN,mvcon,DPGNN,InEXIT} focus on designing complex modeling techniques to model resume-job matching (or referred to as ``person-job fit'').
For example, APJFNN \cite{APJFNN} uses hierarchical recurrent neural networks to process the job and resume content, and DPGNN \cite{DPGNN} uses a dual-perspective graph neural network to model the relationship between resumes and jobs.
However, these methods only show limited improvements, and they often: optimize only for a single task (e.g., interview classification); are hard to accommodate new, unseen resumes or jobs; and are designed for a particular data setting (e.g., applicable only if a specific recruitment platform is used).

In this work, we present a simple method to model resume-job matching, with strong performances in both resume/job ranking \emph{and} resume-job pair classification tasks.
We propose \framework{}, an approach to learn high-quality dense embeddings for resumes and jobs, which can be combined with techniques such as FAISS \cite{FAISS} to rank tens of thousands of resumes and jobs in milliseconds.
To combat label sparsity in person-job fit datasets, \framework{} first uses data augmentation techniques to increase the number of training samples, and then uses contrastive learning \cite{DPR,simlm} to train an encoder.
We evaluate \framework{} on two resume-job matching datasets and find our approach outperforms previous methods (including strong baselines from information retrieval such as BM25) in almost all ranking \emph{and} classification tasks, with up to 20-30\% absolute improvement in ranking jobs and resumes.


\section{Background}
\label{sec:Background}
A resume-job matching (or often called a \emph{person-job fit}) system models the suitability between a resume and a job, allowing it to select the most suitable candidates given a job post, or recommend the most relevant jobs given a candidate's resume \cite{mvcon,DPGNN,InEXIT}.
A job post $J$ (or a resume $R$) is commonly structured as a collection of texts $J = \{\mathbf{x}^{J}_i\}_{i=1}^{p}$, where each piece of text may represent certain sections of the document, such as ``Required Skills'' for a job post or ``Experiences'' for a resume \cite{bian-etal-2019-domain,InEXIT}. Thus, the person-job fit problem is often formulated as a text-based task to model a ``matching'' score:
\[
\mathrm{match}(R, J) = f_\theta(R, J) \to \mathbb{R}
\]
where $f_\theta$ typically involves neural networks \cite{pjfnn,DPGNN,InEXIT} such as BERT \cite{devlin2019bert}.

With the ever-increasing growth of online recruitment data, there is a large number of job posts and resumes (privately) available. However, since a candidate applies to only a small selection of jobs, interactions between resumes and jobs is \emph{very sparse} \cite{mvcon}.
Often, the resulting dataset $\mathcal{D}= \{ R_i, J_i, y_i \}$ has size $|\mathcal{D}| \ll n_{R} \times n_{J}$, where $n_{R}$ and $n_{J}$ are the total number of resumes and jobs respectively, and $y_{i} \in \{0,1\}$ is a \emph{binary} signal representing whether a resume $R_i$ is accepted for an interview by a job $J_i$. While some private datasets \cite{DPGNN} may contain additional labels, such as whether the candidate or recruiter ``requested'' additional information from the other party, we focus on the more common case where only a binary signal is available.

\section{Approach}
\label{sec:ConFit}

We propose \framework{}, a simple and general-purpose approach to model resume-job matching using contrastive learning and data augmentation.
\framework{} produces a dense embedding of a given resume or job post, and models the matching score between an $\langle R,J \rangle$ pair as the inner product of their representations.
This simple formulation allows \framework{} to quickly rank a large number of resumes or jobs when combined with retrieval techniques such as FAISS \cite{FAISS}.
In \Cref{subsec:Data Augmentation}, we describe our approach to augment a person-job fit dataset, and in \Cref{subsec:Contrastive Learning} we describe the contrastive learning approach used during training.


\subsection{Data Augmentation}
\label{subsec:Data Augmentation}

A person-job fit dataset may be considered as a sparse bipartite graph, where each resume $R_i$ and job $J_i$ is a node, and a label (accept or reject) is an edge between the two nodes. Given a resume $R_i$, we first create augmented versions $\hat{R}_i$ by paraphrasing certain sections such as ``Experiences'' (see \Cref{sec:More Details on Data Augmentation} for more details). Since $\hat{R}_i$ includes semantically similar information as $R_i$, we inherit the \emph{same edges} from $R_i$ to $\hat{R}_i$, i.e., any job $J_j$ that accepts $R_i$ for interview also accepts $\hat{R}_i$.
Next, we perform the same augmentation procedure for jobs, creating $\hat{J}_i$ that inherits the same edges as $J_i$ in the graph (which involves augmented $\hat{R}_i$s).
As a result, augmenting $n_{\mathrm{aug}}$ resumes and jobs each \emph{for once} approximately \emph{doubles the number of labeled pairs} (often $\gg n_{\mathrm{aug}}$) in the dataset.
\framework{} thus first performs data augmentation to increase the number of labeled pairs, and then uses contrastive learning (\Cref{subsec:Contrastive Learning}) to train a high-quality encoder. Below, we briefly describe paraphrasing methods used in this work.

\paragraph{EDA Augmentation} Given a piece of text from a resume or a job, we use EDA \cite{wei-zou-2019-eda} to randomly replace, delete, swap, or insert words to create a paraphrased version of the text. We find this to be a simple and fast method to create semantically similar text.

\paragraph{ChatGPT Augmentation} Besides EDA, we also use ChatGPT \cite{chatgpt} to perform paraphrasing. ChatGPT has been used on many data augmentation tasks \cite{chatgpt-paraphrase1,dai2023auggpt}, and in this work, we similarly prompt ChatGPT to paraphrase a given piece of text (see \Cref{sec:More Details on Data Augmentation} for more details).



\subsection{Contrastive Learning}
\label{subsec:Contrastive Learning}

Given an augmented dataset, \framework{} uses contrastive learning \cite{SimCLR,simlm} to further increase the number of training instances from $B$ per batch to $\mathcal{O}(B^2)$ per batch.
Contrastive learning is also an effective technique for learning a high-quality embedding space, and is used in various domains such as information retrieval \cite{DPR} and representation learning \cite{SimCLR,simlm}.

First, we construct contrastive training instances from a dataset $\mathcal{D}= \{ R_i, J_i, y_i \}$:
\[
  \mathcal{D}_\mathrm{con} = \{ \langle R_i^{+}, J_i^{+}, R_{i,1}^{-}, ..., R_{i,l}^{-}, J_{i,1}^{-}, ..., J_{i,l}^{-} \rangle \},
\] 
where each instance contains one positive pair of matched resume-job $\langle R_i^{+}, J_i^{+} \rangle$ with $y_i=1$, and $l$ unsuitable resumes $R_{i,l}^{-}$ for a job $J_i^{+}$ as well as $l$ unsuitable jobs $J_{i,l}^{-}$ for a resume $R_i^{+}$.\footnote{
  Different from information retrieval \cite{DPR,e5} where ranking is an asymmetric task (given a query, rank passages), the person-job fit problem is symmetric (given a resume, rank jobs, and vice versa).
} Following prior work in contrastive learning \cite{SimCLR,gao-etal-2021-simcse,simlm}, we optimize the following cross-entropy loss:
\begin{align}
  &\qquad\qquad \mathcal{L} = \mathcal{L}_{R} + \mathcal{L}_{J} \\
  \mathcal{L}_{R} = &-\log \frac{e^{s_{\theta}(R_i^{+},J_i^{+})}}{e^{s_{\theta}(R_i^{+},J_i^{+})} + \sum_{j=1}^{l} e^{s_{\theta}(R_i^{+},J_{i,j}^{-})}} \nonumber \\
  \mathcal{L}_{J} = &-\log \frac{e^{s_{\theta}(R_i^{+},J_i^{+})}}{e^{s_{\theta}(R_i^{+},J_i^{+})} + \sum_{j=1}^{l} e^{s_{\theta}(R_{i,j}^{-},J_i^{+})}} \nonumber
\end{align}
Similar to training retrieval systems \cite{DPR}, we find the number and choice of negative samples important to obtain a high-quality encoder. We discuss how \framework{} chooses negative samples below.

\paragraph{In-batch negatives} Let there be $B$ positive pairs $\{ \langle R_1^{+}, J_1^{+} \rangle, ... , \langle R_B^{+}, J_B^{+} \rangle\}$ in a mini-batch during training. For each resume $R_i^{+}$, we use the other $B-1$ jobs $\{ J_{j \neq  i}^{+} \}$ as negative samples, and similarly for each job $J_i^{+}$, we use the other $B-1$ resumes as negative samples.
The trick of in-batch negatives thus trains on $B^{2}$ resume-job pairs in each batch, and is highly computationally efficient \cite{gillick-etal-2019-learning,DPR,e5}. In person-job fit, this has a natural interpretation that random (in-batch) negative samples are unsuitable resumes/jobs for a given job/resume.
In practice, we find that using in-batch negatives alone is sufficient to yield competitive ranking performances compared to prior approaches (see \Cref{subsec:Ablation Studies}).

\paragraph{Hard negatives} In addition to in-batch negative samples, we also sample up to $2 \times B_{\mathrm{hard}}$ hard negatives for each batch to further improve \framework{} training. In information retrieval systems, hard negatives \cite{DPR,e5} are often passages that are relevant to the query (e.g. have a high BM25 \cite{bm25} score) but do not contain the correct answer. In person-job fit, we believe that this extends to resumes/jobs that are explicitly \emph{rejected} for a given job/resume. This is because often when a candidate \emph{submits} a resume for a given job post, the resume is \emph{already highly relevant} regardless of whether the candidate is accepted or rejected.
Thus, we sample up to $B_{\mathrm{hard}}$ rejected resumes as hard negatives for any of the $B$ jobs in the mini-batch, as well as $B_{\mathrm{hard}}$ jobs that rejected any of the $B$ resumes. These $2 \times B_{\mathrm{hard}}$ hard negatives are then used by all resumes/jobs in the batch, increasing the number of training pairs to $(B + B_{\mathrm{hard}})^{2} - B_{\mathrm{hard}}^{2}$ per batch.

\subsection{\framework{}}
\label{subsec:Confit_overall}
To address the label sparcity problem in person-job fit datasets, \framework{} first augments the dataset using techniques introduced in \Cref{subsec:Data Augmentation}. Then, \framework{} trains an encoder network $E_\theta$ using contrastive learning described in \Cref{subsec:Contrastive Learning}. Given resumes and job posts during inference, \framework{} first uses the encoder $E_\theta$ to obtain a dense representation for each resume $R$ and job $J$. Then, \framework{} produces a matching score $s_\theta$ between the $\langle R,J \rangle$ pair using inner product:
\[
\mathrm{match}(R,J) = E_\theta(R)^{T} E_\theta(J) \equiv s_{\theta}(R,J)
\]
This simple formulation allows \framework{} to combined with techniques such as FAISS \cite{FAISS} to efficiently rank tens of thousands of resumes and jobs in milliseconds (\Cref{subsec:Runtime}).

\section{Experiments}
\label{sec:Experiments}
We evaluate \framework{} on two real-world person-job fit datasets, and measure its performance and runtime on ranking resumes, ranking jobs, as well as on a fine-grained interview classification task.

\subsection{Dataset and Preprocessing}
\label{subsec:Dataset and Preprocessing}

\paragraph{AliYun Dataset} To our knowledge, the 2019 Alibaba job-resume intelligent matching competition\footnote{
\href{https://tianchi.aliyun.com/competition/entrance/231728/introduction}{https://tianchi.aliyun.com/competition/entrance/231728}
} \emph{provided} the only publicly available person-job fit dataset. All resume and job posts were desensitized and were already parsed into a collection of text fields, such as ``Education'', ``Age'', and ``Work Experiences'' for a resume (see \Cref{sec:More Details on Dataset and Preprocessing} for more details). All resumes and jobs are in Chinese.

\paragraph{Intellipro Dataset} The resumes and job posts are collected from a global hiring solution company, called ``Intellipro Group Inc.''\ref{author:intellipro}. To protect the privacy of candidates, all records have been anonymized by removing sensitive identity information.
For each resume-job pair, we record whether the candidate is accepted ($y=1$) or rejected ($y=0$) for an interview. For generalizability, we parse all resumes and jobs into similar sections/fields as the AliYun dataset. Both English and Chinese resumes and jobs are included.

Since neither dataset has an official test set, we first construct test sets with statistics shown in \Cref{tbl:test_dset}. To measure the ranking ability of current methods, we consider two tasks: 1) ranking $q=100$ resumes given a job post (denoted as \emph{Rank Resume}), and 2) ranking $q=100$ jobs given a resume (denoted as \emph{Rank Job}).
Since only a few resumes and jobs are labeled, we fill in random resumes/jobs to reach $q$ slots when needed.
We further consider the ``fine-grained'' scoring ability of current methods, by measuring how well a method can distinguish between an accepted resume-job pair and a rejected one (denoted as \emph{classification}). We exclude all resumes and jobs used in test and validation sets from the training set, and present the training, test, and validation set statistics in \Cref{tbl:train_dset}, \Cref{tbl:test_dset} and \Cref{tbl:val_dset}, respectively.
\begin{table}[!t]
  \centering
  \scalebox{0.8}{
    \begin{tabular}{l cc}
      \toprule
      Train & \textbf{Intellipro Dataset} & \textbf{Aliyun Dataset}\\
      \midrule
      \# Jobs     & 1794 & 19542 \\
      \# Resumes  & 6435 & 2718 \\
      \# Labels   & 6751 & 22124 \\
      \phantom{--}\textcolor{gray}{(\# accept)} & 2809 & 10185 \\
      \phantom{--}\textcolor{gray}{(\# reject)} & 3942 & 11939 \\
      \# Industries & 16 & 20 \\  
      \cmidrule(lr){1-3}
      \# Fields per $R$ & 8 & 12 \\
      \# Fields per $J$ & 9 & 11 \\
      \cmidrule(lr){1-3}
      \# Words per $R$ & 915.2 & 101.9 \\
      \# Words per $J$ & 174.7 & 153.1 \\
      \bottomrule
    \end{tabular}
  }
\caption{Training dataset statistics. \emph{\# Words per $R$/$J$} represent the \emph{average} number of words per resume/job.}
\label{tbl:train_dset}
\end{table}
\begin{table}[!t]
  \centering
  \scalebox{0.63}{
    \begin{tabular}{l ccc ccc}
      \toprule
      & \multicolumn{3}{c}{\textbf{Intellipro Dataset}} & \multicolumn{3}{c}{\textbf{AliYun Dataset}}\\
      \textbf{Test} & Rank $R$ & Rank $J$ & Classify & Rank $R$ & Rank $J$ & Classify \\
      \midrule
      \# Samples
      & 120 & 120 & 120 
      & 300 & 300 & 300 \\
      \# Jobs
      & 120 & 427 & 104
      & 300 & 2903 & 299 \\
      \# Resumes
      & 1154 & 120 & 117 
      & 1006 & 300 & 280\\
      \bottomrule
    \end{tabular}
  }
\caption{Test dataset statistics. \emph{Classify} is a binary classification task to predict whether a resume-job pair is accepted or rejected for interview.}
\vspace{-10pt}
\label{tbl:test_dset}
\end{table}
\subsection{Model Architecture}
\label{subsec:Model Architecture}

\begin{figure}[t!]
    \centering
    \includegraphics[scale=0.88]{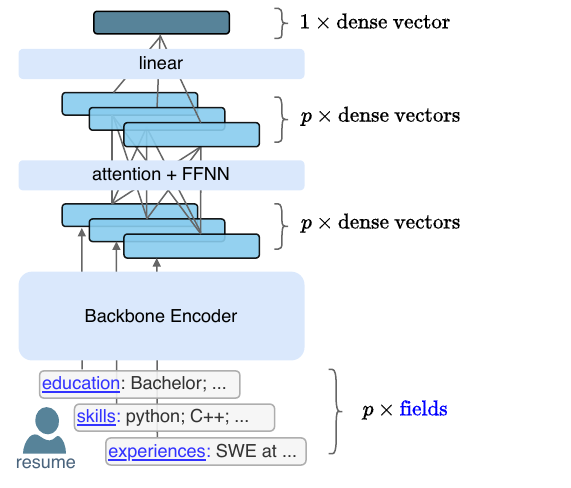}
    \caption{Model architecture used to encode a resume or a job post, formatted as a collection of $p$ text fields (see \Cref{sec:More Details on Dataset and Preprocessing} for a full example of resume/job).
    }
    \label{fig:model_archi}
    \vspace{-8pt}
\end{figure}

Since both datasets represent resumes and job posts as a collection of text fields, we simplify the model architecture from InEXIT \cite{InEXIT}, outlined in \Cref{fig:model_archi}.
InEXIT encodes each text field (e.g., ``education: Bachelor;...'') in a resume or a job independently using a pre-trained encoder, and considers a hierarchical attention mechanism to model person-job fit as interactions between these fields. Following InEXIT, we first encode each field independently, and model the ``internal interaction'' between the fields \emph{within} a resume/job using attention \cite{vaswani2023attention}.
InEXIT then uses another attention layer on all text fields of the resume-job pair to model the ``external interaction'' \emph{between} a resume and a job post, and finally produces a score using an MLP layer (see \Cref{sec:more_details_on_model_architecture} for more details).
Since \framework{} models person-job fit based on \emph{independently} produced resume/job embeddings, we replace the last attention and MLP layer with a linear layer, which directly fuses the field representations into a single dense vector for a given resume or a job post (\Cref{fig:model_archi}).
%
\begin{table*}[!t]
  \centering
  \scalebox{0.68}{
    \begin{tabular}{ll ccc cccc ccc cccc}
      \toprule
      & & \multicolumn{7}{c}{\textbf{Intellipro Dataset}} & \multicolumn{7}{c}{\textbf{AliYun Dataset}} \\
      & & \multicolumn{2}{c}{Rank Resume} & \multicolumn{2}{c}{Rank Job} & \multicolumn{3}{c}{Classification} 
      & \multicolumn{2}{c}{Rank Resume} & \multicolumn{2}{c}{Rank Job} & \multicolumn{3}{c}{Classification} \\
      \cmidrule(lr){3-4} \cmidrule(lr){5-6} \cmidrule(lr){7-9} 
      \cmidrule(lr){10-11} \cmidrule(lr){12-13} \cmidrule(lr){14-16}
      \textbf{Method} & \textbf{Encoder} 
      & MAP & nDCG & MAP & nDCG & F1 & Prc+ & Rcl+ 
      & MAP & nDCG & MAP & nDCG & F1 & Prc+ & Rcl+ \\
      \midrule
      \multirow{3}{*}{XGBoost}
      & BoW 
      & 14.42 & 11.84 & 3.89 & 2.98
      & 49.51 & 37.74 & 41.67 
      & 6.67 & 6.25 &13.63 & 12.53
      & 69.57 & 72.00 & 54.14 \\
      & TF-IDF
      & 10.18 & 7.87  & 4.11 & 2.94
      & 55.00 & 43.75 & 43.75 
      & 5.56  & 4.96  &13.85 & 13.21
      & 68.78 & 69.16 & 55.64 \\
      & BERT-base
      & 8.45  & 7.69  & 5.57 & 5.24
      & 61.01 & 50.98 & \textbf{54.17}
      & 5.34 & 5.10 &13.45 &12.22
      & 70.63 & \textbf{73.27} & 55.64 \\
      \cmidrule(lr){2-16}
      \multirow{2}{*}{RawEmbed.}
      & E5-small
      & 28.61 & 33.88 & 25.48 & 30.26
      & 54.20 & 42.86 & 31.25 
      & 16.06 & 17.89 & 20.26 & 22.84
      & 38.64 & 27.78 & 22.56 \\
      & BERT-base
      & 13.07 & 13.94 & 4.41 & 3.62
      & 49.25 & 34.38 & 22.92
      & 9.18  & 10.63 &12.35 &12.63
      & 46.71 & 40.00 & 40.60 \\
      \cmidrule(lr){2-16}
      MV-CoN
      & BERT-base
      & 10.81 & 10.00 & 3.34 & 2.17
      & 58.00 & 50.00 & 33.33 
      & 5.41  & 5.15  & 13.44 & 12.67
      & \textbf{74.25} & \second{\textbf{72.22}} & \second{\textbf{68.32}}\\
      InEXIT
      & BERT-base
      & 12.27 & 12.98 & 4.11 & 3.46
      & 55.55 & 44.74 & 35.42 
      & 5.25  & 4.98  & 13.02 & 12.30
      & \second{\textbf{71.75}} & 66.67 & \textbf{72.18} \\
      DPGNN
      & BERT-base
      & 19.64 & 21.95 & 17.86 & 19.60
      & \second{\textbf{61.16}} & \second{\textbf{52.38}} & 45.83 
      & 19.96 & 24.64 & 27.23 & 30.07
      & 50.31 & 45.24 & 57.14 \\
      BM25
      & -
      & \second{\textbf{39.13}} & \second{\textbf{44.96}} & \second{\textbf{37.88}} & \second{\textbf{43.15}}
      & - & - & - 
      & \textbf{34.71} & \textbf{40.56} & \second{\textbf{27.30}} & \second{\textbf{31.18}} 
      & - & - & - \\
      \cmidrule(lr){2-16}
      Ours
      & BERT-base
      & {\textbf{44.47}} & {\textbf{49.51}}  
      & {\textbf{39.57}} & {\textbf{45.67}}
      & {\textbf{63.78}} & {\textbf{55.81}} & \second{\textbf{50.00}}
      & \second{\textbf{30.79}} & \second{\textbf{37.71}}  
      & {\textbf{36.13}} & {\textbf{41.65}}
      & 47.16 & 41.10 & 45.11\\
      \bottomrule
    \end{tabular}
  }
  \caption{Comparing ranking and classification performance of various approaches when a small encoder is used. \emph{F1} is weighted F1 score, \emph{nDCG} is nDCG@10, \emph{Prc+} and \emph{Rcl+} are precision and recall for positive classes. Results for non-deterministic methods are averaged over 3 runs. Best result is shown in \textbf{bold}, and runner-up is in \second{\textbf{gray}}.}
  \label{tbl:main_exp_bert}
  \vspace{-5pt}
\end{table*}
\begin{table*}[!t]
\centering
\scalebox{0.68}{
  \begin{tabular}{ll ccc cccc ccc cccc}
    \toprule
    & & \multicolumn{7}{c}{\textbf{Intellipro Dataset}} & \multicolumn{7}{c}{\textbf{AliYun Dataset}} \\
    & & \multicolumn{2}{c}{Rank Resume} & \multicolumn{2}{c}{Rank Job} & \multicolumn{3}{c}{Classification} 
    & \multicolumn{2}{c}{Rank Resume} & \multicolumn{2}{c}{Rank Job} & \multicolumn{3}{c}{Classification}\\
    \cmidrule(lr){3-4} \cmidrule(lr){5-6} \cmidrule(lr){7-9} 
    \cmidrule(lr){10-11} \cmidrule(lr){12-13} \cmidrule(lr){14-16}
    \textbf{Method} & \textbf{Encoder} 
    & MAP & nDCG & MAP & nDCG & F1 & Prc+ & Rcl+ 
    & MAP & nDCG & MAP & nDCG & F1 & Prc+ & Rcl+ \\
    \midrule
    \multirow{3}{*}{XGBoost}
    & BoW 
    & 14.42 & 11.84 & 3.89 & 2.98 
    & 49.51 & 37.74 & 41.67 
    & 6.67  & 6.25  &13.63 &12.53
    & 69.57 & 72.00 & 54.14 \\
    & TF-IDF
    & 10.18 & 7.87  & 4.11  & 2.94
    & 55.00 & 43.75 & 43.75 
    & 5.56  & 4.96  & 13.85 &13.21
    & 68.78 & 69.16 & \second{\textbf{55.64}} \\
    & text-ada-002
    & 9.87  & 9.94  & 4.15 & 3.58
    & \textbf{62.43} & \textbf{53.19} & \second{\textbf{52.08}}
    & 6.40  & 6.08  &13.46 &12.93
    & 62.43 & 53.19 & 52.08 \\
    \cmidrule(lr){2-16}
    \multirow{3}{*}{RawEmbed.}
    & xlm-roberta-l
    & 14.46 & 14.95 & 13.22 & 13.94
    & 51.27 & 40.00 & 16.67 
    & 7.07  & 6.90  & 11.25 & 10.75
    & 53.48 & 47.62 & 52.63 \\
    & E5-large
    & 35.10 & 40.10 & 26.93 & 30.61
    & \second{\textbf{59.39}} & \second{\textbf{51.43}} & 37.50 
    & 32.11 & 37.45 & 24.56 & 28.15
    & 44.67 & 35.64 & 27.07 \\
    & text-ada-002
    & \second{\textbf{42.85}} & \second{\textbf{48.11}} & \textbf{43.28} & \second{\textbf{51.11}}
    & 58.92 & 48.89 & 45.83 
    & 31.47 & 37.06 & 21.94 & 24.80
    & 39.72 & 32.35 & 33.08 \\
    \cmidrule(lr){2-16}
    MV-CoN
    & E5-large
    & 12.23 & 10.98 & 4.06  & 3.09
    & 52.75 & 40.51 & 27.08 
    & 5.60  & 5.15 & 12.92 & 12.67 
    & \textbf{70.85} & \textbf{77.53} & 51.88 \\
    InEXIT
    & E5-large
    & 13.60 & 13.63 & 3.14 & 1.91
    & 55.17 & 45.16 & 29.17
    & 5.49  & 4.33 & 13.39 & 13.21 
    & \second{\textbf{70.41}} & \second{\textbf{74.74}} & 53.58 \\
    DPGNN
    & E5-large
    & 21.31 & 24.08 & 13.90 & 17.69
    & 55.56 & 48.00 & 25.00
    & 33.98 & 40.63 & \second{\textbf{42.76}} & \second{\textbf{46.98}}
    & 53.88 & 48.03 & 45.86 \\
    BM25
    & -
    & 39.13 & 44.96 & 37.88 & 43.15
    & - & - & - 
    & \second{\textbf{34.71}} & \second{\textbf{40.56}} & 27.30 & 31.18
    & - & - & - \\
    \cmidrule(lr){2-16}
    Ours
    & E5-large
    & \textbf{43.08} & \textbf{50.28} & \second{\textbf{42.88}} & \textbf{51.74}
    & 54.09 & 45.21 & \textbf{68.75}
    & \textbf{64.32} & \textbf{71.71} & \textbf{59.13} & \textbf{65.91}
    & 59.71 & 53.66 & \textbf{66.17} \\
    \bottomrule
  \end{tabular}
}
\caption{Comparing ranking and classification performance of various approaches when a larger encoder is used. \emph{F1} is weighted F1 score, \emph{nDCG} is nDCG@10, \emph{Prc+} and \emph{Rcl+} are precision and recall for positive classes. Results for non-deterministic methods are averaged over 3 runs. Best result is shown in \textbf{bold}, and runner-up is in \second{\textbf{gray}}.}
\label{tbl:main_exp_e5}
\vspace{-10pt}
\end{table*}
\subsection{Baselines}
\label{subsec:Baselines}

We compare \framework{} against both recent best person-job fit systems and strong baselines from information retrieval systems.

Recent person-job fit systems can be grouped into two categories: classification-targeted and ranking-targeted. The \textbf{best} classification-targeted system include \emph{MV-CoN} \cite{mvcon} and \emph{InEXIT} \cite{InEXIT}.
MV-CoN considers a co-teaching network \cite{han2018coteaching} to learn from sparse, noisy person-job fit data, and InEXIT uses hierarchical attention to model interactions between the text fields of a resume-job pair.
Both methods optimize for the classification task.
The \textbf{best} ranking-targeted systems include \emph{DPGNN} \cite{DPGNN}. DPGNN considers a dual-perspective graph view of person-job fit and uses a BPR loss \cite{rendle2012bpr} to optimize for resume and job ranking.

We also compare against methods from information retrieval systems such as: \emph{BM25} \cite{bm25,bm25-all} and \emph{RawEmbed}. BM25 is a strong baseline used for many text ranking tasks \cite{thakur2021beir,e5,kamalloo2023beir-resources}, and RawEmbed is based on dense retrieval methods \cite{DPR,FAISS} that directly concatenates all text fields and uses a pre-trained encoder to produce a single dense embedding for inner product scoring.
Finally, we also consider \emph{XGBoost} \cite{XGBoost} as a generic method for classification and ranking tasks, where features can be Bag-of-Words (BoW), TF-IDF vectors, and pre-trained embeddings from RawEmbed.

Unless otherwise indicated, \framework{} first uses data augmentation with both EDA and ChatGPT, each augmenting 500 resumes and 500 jobs for each dataset (\Cref{subsec:Data Augmentation}), followed by contrastive learning with $B=8$ and $B_{\mathrm{hard}}=8$ (\Cref{subsec:Contrastive Learning}). See \Cref{sec:Training Hyperparameters} for other hyperparameters used by \framework{}, and see \Cref{sec:more_details_on_baselines} for more implementation details of the baselines.
\subsection{Metrics}
\label{subsec:Metrics}
Following prior work \cite{DPR,DPGNN}, we use Mean Average Precision (MAP) and normalized Discounted Cumulative Gain (nDCG) to measure the ranking ability of each method. Since most resume-job pairs are unlabeled, we report nDCG@10. To measure the fine-grained classification ability of a method, we follow prior work in person-job fit \cite{APJFNN,pjfnn,mvcon,InEXIT} and use weighted F1, precision, and recall.
Since correctly predicting a positive sample (i.e., a suitable job for a resume) is important in practice, we report precision and recall for the positive class (denoted as \emph{Prc+} and \emph{Rcl+}, respectively).

\subsection{Main Results}
\label{subsec:Main Results}
\Cref{tbl:main_exp_bert} summarizes \framework{}'s performance in comparison to other baselines, when an encoder with $\sim$180M parameters is used as the backbone. This includes using BERT-base\footnote{Since the AliYun dataset is solely in Chinese, we use BERT-base-chinese for the AliYun dataset and BERT-base-multilingual-cased for the Intellipro dataset.} \cite{devlin2019bert} and E5-small \cite{e5}.
In general, we find that classification-targeted systems such as \emph{MV-CoN} and \emph{InEXIT} achieve a high F1 score but have poor ranking ability, while ranking-targeted methods such as \emph{RawEmbed}, \emph{DPGNN}, and \emph{BM25} perform much better in ranking.
With a small encoder model, we find \framework{} achieves the best ranking performance in three out of the four tasks, and BM25 achieves the best in the remaining task. \framework{} also achieves the best F1 score on the Intellipro classification task compared to other classification-targeted systems.

\Cref{tbl:main_exp_e5} summarizes each method's performance when a larger backbone encoder ($\sim$560M parameters) is used. This includes multilingual-E5-large \cite{e5}, xlm-roberta-large \cite{xlm-roberta,liu2019roberta}, or OpenAI text-ada-002\footnote{Model size unknown.} \cite{text-ada}.
Similar to \Cref{tbl:main_exp_bert}, we find that classification-targeted methods such as \emph{MV-CoN} reach a high F1 score, while ranking-targeted methods achieve a better MAP and nDCG score.
We also find that \framework{} now achieves the best ranking performances in all cases, except for the MAP score in the IntelliPro's job ranking task.
We believe this is because the IntelliPro dataset contains much less data compared to the AliYun dataset (\Cref{tbl:train_dset}).
In the AliYun dataset, \framework{} improves up to $\sim$30\% absolute in MAP and nDCG score for ranking resumes and up to $\sim$20\% for ranking jobs.
We believe this is because the AliYun dataset not only has more data, but also uses much shorter and concise texts compared to the Intellipro dataset (\Cref{tbl:train_dset}).
Lastly, we find \framework{} remains competitive in classification task for both datasets, despite not directly optimizing for them.
%
\subsection{Runtime Analysis}
\label{subsec:Runtime}
\begin{figure}[t!]
  \centering
  \includegraphics[scale=0.45]{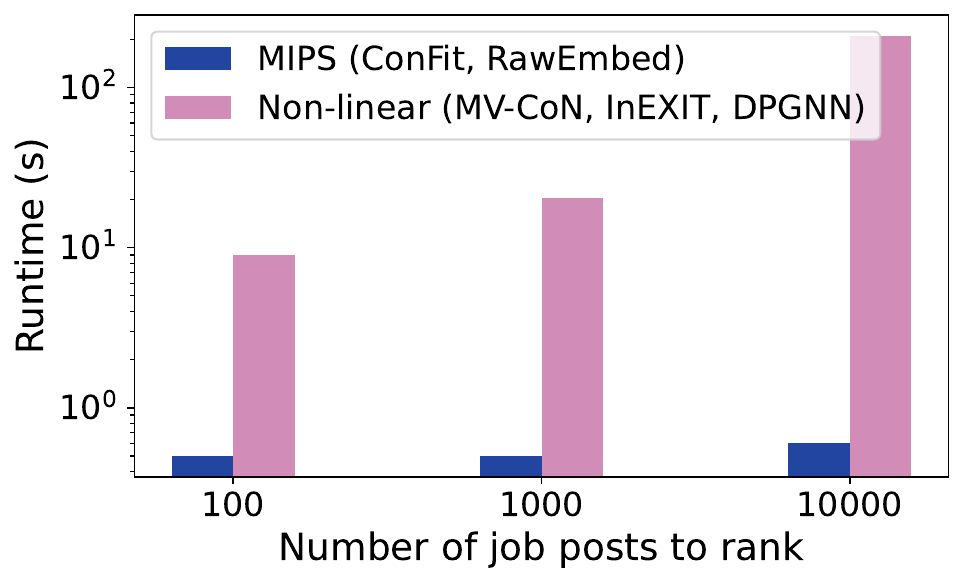}
  \caption{Runtime comparison between neural-based methods. \emph{MIPS} are maximum inner product search methods that are supported by FAISS \cite{FAISS}. \emph{Non-linear} methods require an additional forward pass to produce a score between a resume-job pair. Results are averages over three runs.}
  \label{fig:speed}
  \vspace{-4pt}
\end{figure}

A practical recruitment system needs to \emph{quickly} rank a large number of resumes given a job post, or vice versa.
We measure the runtime to rank 100; 1,000; and 10,000 jobs for a given resume from the AliYun dataset, and compare the speed of various neural-based methods from \Cref{tbl:main_exp_bert}.
We present the results in \Cref{fig:speed}.
In general, methods that ranks by inner product search (\emph{RawEmbed} and \framework{}) can utilize FAISS \cite{FAISS} to achieve a runtime in milliseconds in all cases\footnote{After embedding all relevant resume and job posts, which only needs to be computed \emph{once}.}.
However, methods such as \emph{MV-CoN}, \emph{InEXIT}, and \emph{DPGNN} requires a (partial) forward pass for each resume-job pair to produce a score between (see \Cref{sec:More Details on Runtime Comparison} for more details).
We believe this is highly inefficient, especially when the number of documents to rank (e.g., job posts) is large.
%
\begin{table*}[t!]
\centering
\scalebox{0.73}{
  \begin{tabular}{l ccc cccc ccc cccc}
    \toprule
    & \multicolumn{7}{c}{\textbf{Intellipro Dataset}} & \multicolumn{7}{c}{\textbf{AliYun Dataset}} \\
    & \multicolumn{2}{c}{Rank Resume} & \multicolumn{2}{c}{Rank Job} & \multicolumn{3}{c}{Classification} 
    & \multicolumn{2}{c}{Rank Resume} & \multicolumn{2}{c}{Rank Job} & \multicolumn{3}{c}{Classification}\\
    \cmidrule(lr){2-3} \cmidrule(lr){4-5} \cmidrule(lr){6-8} 
    \cmidrule(lr){9-10} \cmidrule(lr){11-12} \cmidrule(lr){13-15}
    \textbf{Modification} 
    & MAP & nDCG & MAP & nDCG & F1 & Prc+ & Rcl+ 
    & MAP & nDCG & MAP & nDCG & F1 & Prc+ & Rcl+ \\
    \midrule
      +contrastive &
      \bestunderline{42.96} & \bestunderline{49.28} & \bestunderline{42.23} & \bestunderline{49.21} &
      59.17 & 56.00 & 29.17 &
      \bestunderline{27.53} & \bestunderline{33.34} & \bestunderline{34.05} & \bestunderline{37.83} &
      \bestunderline{47.60} & \bestunderline{40.77} & 39.85 \\
      \phantom{--}\textcolor{gray}{$B_{\mathrm{hard}}=0$} &
      31.15 & 36.88 & 38.94 & 44.89 &
      52.02 & 44.16 & \bestunderline{70.83} &
      13.73 & 15.43 & 32.52 & 37.23 & 
      43.31 & 34.51 & 29.32\\
      \phantom{--}\textcolor{gray}{$B_{\mathrm{hard}}=2$} &
      41.85 & 47.77 & 41.24 & 47.46 &
      56.14 & 56.58 & 19.79 &
      25.42 & 31.08 & 31.59 & 35.64 &
      45.65 & 38.64 & 38.35\\
      \phantom{--}\textcolor{gray}{$B_{\mathrm{hard}}=4$} &
      40.86 & 45.69 & 41.29 & 47.78 &
      \bestunderline{61.69} & \bestunderline{62.36} & 33.33 &
      25.60 & 30.65 & 32.72 & 36.15 &
      44.51 & 38.67 & \bestunderline{43.61} \\
    \cmidrule(lr){2-15}
    +Data Aug. &
    \bestunderline{44.47} & \bestunderline{49.51} & 39.57 & 45.67 &
    \bestunderline{63.78} & 55.81 & \bestunderline{50.00} &
    \bestunderline{30.79} & \bestunderline{37.71} & \bestunderline{36.13} & \bestunderline{41.65} &
    \bestunderline{47.16} & \bestunderline{41.10} & \bestunderline{45.11}\\
    \phantom{--}\textcolor{gray}{ChatGPT only} &
    41.36 & 47.45& 39.54 & \bestunderline{46.98} &
    58.75 & \bestunderline{71.43} & 20.83 &
    29.69 & 35.59 & 35.11 & 40.13 &
    47.00 & 39.67 & 36.09 \\
    \phantom{--}\textcolor{gray}{EDA only} &
    39.49 & 46.03 & \bestunderline{40.25} & 46.03 & 60.64 & 65.00 & 27.08 &
    27.06 & 33.08 & 34.69 & 39.16 & 46.30 & 38.39 & 32.33 \\
    \phantom{--}\textcolor{gray}{EDA-all} &
    39.03 & 45.56 & 40.09 & 45.76 &
    60.64 & 65.00 & 27.08 &
    28.21 & 33.77 & 33.77 & 38.13 &
    45.59 & 37.82 & 33.83 \\
    \bottomrule
  \end{tabular}
}
\caption{\framework{} ablation studies. \framework{} uses contrastive learning (\emph{+contrastive}) with $B_{\mathrm{hard}}=8$, and Data Augmentation (\emph{+Data Aug.}) with both ChatGPT and EDA.
Best result in each ablation group is highlighted in \bestunderline{bold}.
}
\vspace{-12pt}
\label{tbl:ablation_studies}
\end{table*}
\subsection{Ablation Studies}
\label{subsec:Ablation Studies}
\Cref{tbl:ablation_studies} presents our ablation studies for each component of \framework{} training.
We focus on using BERT-base from \Cref{tbl:main_exp_bert} as it is less resource-intensive to train.

First, we consider \framework{} to only use contrastive learning (denoted as \emph{+contrastive}) under various settings, such as $B=8,B_{\mathrm{hard}}=\{0,2,4,8\}$.
In \Cref{tbl:ablation_studies}, we find that: a) increasing the number of hard negatives ($B_{\mathrm{hard}}$) improves ranking performance, and b) using contrastive learning alone already outperforms many baselines in \Cref{tbl:main_exp_bert}.
This suggests that contrastive learning plays a major role in \framework{}'s performance.

Next, we add data augmentation to training, and measure the performance of: 1) using only ChatGPT to augment 500 resumes and jobs, denoted as \emph{ChatGPT only}; 2) using EDA to augment 500 resumes and jobs, denoted \emph{EDA only}; 3) using EDA to augment all resume/job seen during training, denoted as \emph{EDA-all}; and 4) combining both 1) and 2), denoted as \emph{+Data Aug.}
In general, we find combining both ChatGPT and EDA augmentation can most often achieve the best performance. 
We believe this is because such approach includes both \emph{semantically paraphrased} content from ChatGPT and \emph{syntactically altered} content (e.g., inserting or removing words) from EDA.
Especially for the AliYun dataset, we find using any form of data augmentation improves over using contrastive learning alone.
We believe this is because AliYun's resume/job texts are much shorter and more concise than those of the Intellipro dataset, thus making data augmentation easier to perform.

Since \framework{} training is model-agnostic, we also experiment with \emph{completely removing neural networks}, and only use TF-IDF representations with XGBoost. 
Despite seeing performance degradation compared to \framework{} with pretrained encoders, we find this approach is \emph{still competitive} against prior best person-job fit systems that uses BERT (see \Cref{sec:More Details on Ablation Studies} and \Cref{tbl:main_exp_xgboost} for more details). This suggests that the contrastive learning and data augmentation \emph{procedure} from \framework{} is effective for the person-job fit task.
\begin{figure}[t!]
  \centering
  \includegraphics[scale=0.32]{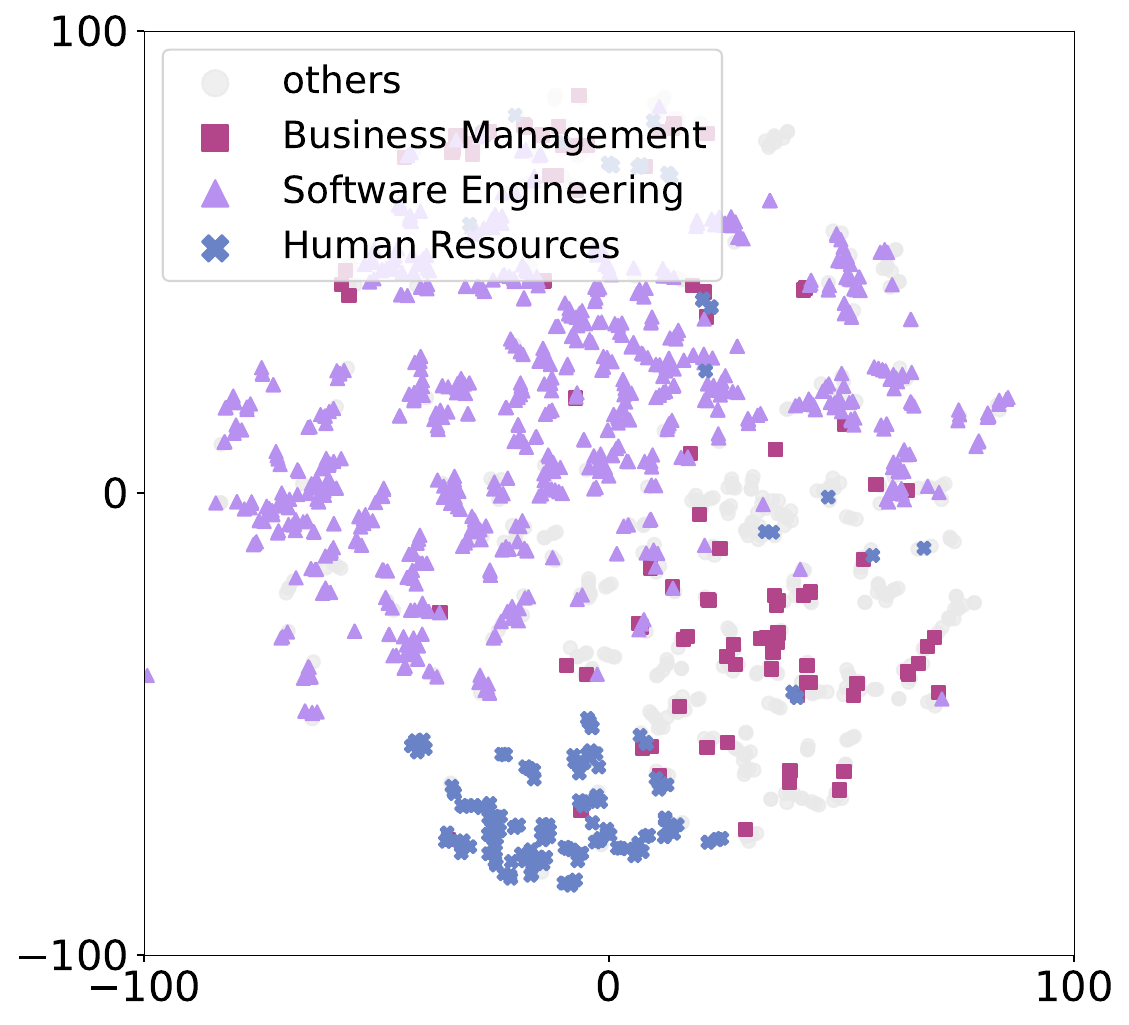}
  \caption{Visualizing resume embeddings from \framework{} using t-SNE. Colors are assigned using each resume's desired industry. Top-3 most frequent industries are color-coded for easier viewing.}
  \label{fig:confit_resume_embedding}
  \vspace{-12pt}
\end{figure}
\section{Analysis}
\label{sec:Analysis}
In this section, we provide both qualitative visualization and quantitative analysis of the embeddings learned by \framework{}. We mainly focus on the Intellipro dataset as it is more challenging.

\subsection{Qualitative Analysis}
\label{subsec:Qualitative Analysis}

\framework{} aims to learn a high-quality embedding space for a resume or a job post. In \Cref{fig:confit_resume_embedding} we visualize the resume embeddings learned by \framework{}.
We use \framework{} with BERT-base (see \Cref{tbl:main_exp_bert}) to embed all 1457 resumes from the test set in the Intellipro dataset, and perform dimensionality reduction using t-SNE \cite{tsne}.
In \Cref{fig:confit_resume_embedding}, we find \framework{} learned to cluster resumes based on important fields such as ``Desired Industry''. We believe this is consistent with how a human would determine person-job fit, as resumes aiming for similar industries are likely to contain similar sets of experiences and skills.
For comparison with embeddings generated by other baselines, please see \Cref{sec:More Qualitative Analysis}.
\subsection{Error Analysis}
\label{subsec:Error Analysis}
To analyze the errors made by \framework{}, we manually inspect 30 \emph{negative} resume-job pairs from the ranking tasks that are \emph{incorrectly ranked at top 5\%} and is before at least one positive pair, and 20 pairs from the classification task that was incorrectly predicted as a match.
For each incorrectly ranked or classified pair, we compare against other positive resume-job pairs from the dataset, and categorize the errors with the following criteria: \emph{unsuitable}, where some requirements in the job post are not satisfied by the resume; \emph{less competent}, where a resume satisfies all job requirements, but many {competing candidates} have a higher degree/more experience; \emph{out-of-scope}, where a resume satisfies all requirements, appears competitive compared to other candidates, but is still rejected due to other (e.g., subjective) reasons not presented in our resume/job data themselves; and \emph{potentially suitable}, where a resume from the ranking tasks satisfied the requirements and seemed competent, but had no label in the original dataset.

\begin{figure}[t!]
  \centering
  \includegraphics[scale=0.32]{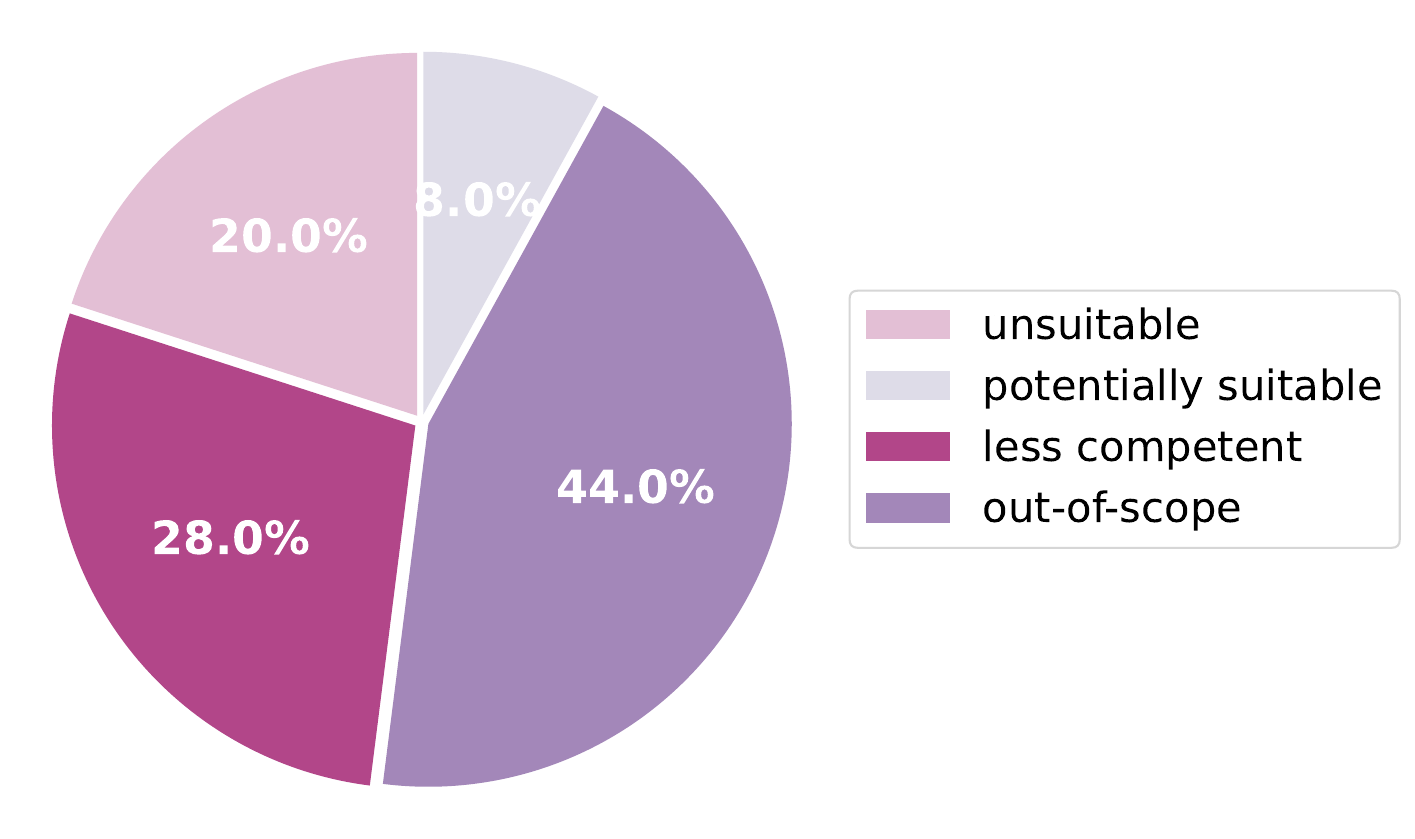}
  \caption{\framework{} error analysis. We find 44\% of the errors made are due to reasons not identifiable using resume/job documents alone, and 28\% due to a candidate's resume satisfying all the job requirements but is less competent than \emph{other competing candidates}.}
  \label{fig:error_frequencies}
\end{figure}

We present our analysis in \Cref{fig:error_frequencies}, and find that a significant portion of errors are \emph{out-of-scope}, where we believe information in resumes/job posts is limited. The next most frequent error is \emph{less competent}, which is understandable since \framework{} scores a resume-job pair independent of other competing candidates. Lastly, we also find that about 20\% of the wrong predictions were \emph{unsuitable}, with resumes not satisfying certain job requirements such as ``4 years+ with Docker, K8s''.
We believe \emph{unsuitable} errors may be mitigated by combining \framework{} with better feature engineering techniques along with keyword-based approaches (such as BM25), which we leave for future work.
\section{Related Work}
\label{sec:Related Work}
\paragraph{Person-job fit systems} Early neural-based methods in person-job fit \cite{GUO2016169} typically focus on network architecture to obtain a good representation of a job post or a resume. These methods include \citet{APJFNN,pjfnn,cnn-lstm-siamese,jiang2020learning,10169716}, which explores
architectures such as RNN, LSTM \cite{staudemeyer2019understanding} and CNN \cite{oshea2015introduction}. Recent deep learning methods include \citet{siamese,cnn-lstm-siamese}, which uses deep siamese network to learn an embedding space for resume/jobs, \citet{bian-etal-2019-domain} which uses a hierarchical RNN to improve domain-adaptation of person-job fit systems, and \citet{Zhang2023FedPJFFC} which uses federated learning to perform model training while preserving user privacy. However, as person-job fit systems involve sensitive data, most systems do not open-source datasets \emph{or implementations}, and are often optimized for one particular dataset. Recent work with public implementations includes MV-CoN \cite{mvcon}, which uses a co-teaching network \cite{malach2018decoupling} to perform gradient updates based model's confidence to data noises; InEXIT \cite{InEXIT}, which uses hierarchical attention to model resume-job interactions; and DPGNN \cite{DPGNN}, which uses a graph-based approach with a novel BPR loss to optimize for resume/job ranking. \framework{} uses contrastive learning and data augmentation techniques based on powerful pre-trained models such as BERT \cite{devlin2019bert}, and achieves the best performance in almost all ranking and classification tasks across two person-job fit datasets.

\paragraph{Information retrieval systems} \framework{} benefits from contrastive learning techniques, which have seen wide applications in many information retrieval and representation learning tasks \cite{SimCLR,radford2021learning}. Given a query (e.g., user-generated question), an information retrieval system aims to find top-k relevant passages from a large reserve of candidate passages \cite{joshi2017triviaqa,natural-questions}.
Popular methods in information retrieval include BM25 \cite{bm25,bm25-all}, a keyword-based approach used as the baseline in many text ranking tasks \cite{ms-marco,thakur2021beir,muennighoff2022mteb}, and dense retrieval methods such as \citet{DPR,izacard2021contriever,e5}, which uses contrastive learning to obtain high-quality passage embeddings and typically performs top-k search based on inner product. To our knowledge, \framework{} is the first attempt to use contrastive learning for person-job fit, achieving the best performances in almost all person-job ranking tasks across two different person-job fit datasets.

\section{Conclusion}
\label{sec:Conclusion}
We propose \framework{}, a general-purpose approach to model person-job fit.
\framework{} trains a neural network using contrastive learning to obtain a high-quality embedding space for resumes and job posts, and uses data augmentation to alleviate data sparsity in person-job fit datasets.
Our experiments across two person-job fit datasets show that \framework{} achieves the best performance in almost all ranking and classification tasks.
We believe \framework{} is easily extensible, and can be used as a strong foundation for future research on person-job fit.

\section{Limitations}
\label{sec:Limitations}

\paragraph{Recruiter/Job Seeker Preference} \framework{} produces dense representations for resumes and jobs \emph{independently}, and uses inner-product to score the resume-job pair. While this approach can be easily combined with retrieval methods such as FAISS \cite{FAISS} to efficiently rank a large number of resumes/jobs, it ignores certain aspects of how a real recruiter or a job seeker may choose a resume or a job.
In our error analysis (\Cref{subsec:Error Analysis}), we find a significant portion of incorrectly ranked/rated resume-job pairs \emph{could} be either due to subjective choices made by the recruiters, or due to a very competitive candidate pool for a certain job position.
This suggests that additionally modeling the recruiter or job seeker's past preferences (e.g., using profiling approaches \cite{profiling-choice} from recommendation systems \cite{eliyas2022recommendation}) may be beneficial, and that developing a scoring metric that is \emph{aware of the other candidates} in the pool could also be useful.
In general, we believe \framework{} embeddings would serve as a foundation for these approaches, and we leave this for future work.

\paragraph{Sensitive Data} To our knowledge, there is no standardized, public person-job fit dataset\footnote{The AliYun dataset used in this work is no longer publicly available as of 09-11-2023.} that can be used to compare performances of existing systems \cite{pjfnn,APJFNN,mvcon,DPGNN,InEXIT}.
This is understandable, as resume contents contain highly sensitive information and that large-scale person-job datasets are often proprietary.
We provide our best effort to make \framework{} reproducible and extensible for future work: we will open-source full implementations of \framework{} and all relevant baselines, our data processing scripts, and dummy train/valid/test data files that can be used test drive our system end-to-end. We will also privately release our model weights and full datasets to researchers under appropriate license agreements.
We hope these attempts can make future research in person-job fit more accessible.

\section{Ethical Considerations}
\label{sec:Ethical Considerations}

\framework{} uses pretrained encoders such as BERT and E5 \cite{devlin2019bert,e5}, and it is well-known that many powerful encoders contain biases \cite{pmlr-v97-brunet19a,social-bias,jentzsch-turan-2022-gender,Caliskan_2022}. For person-job fit systems, we believe it is crucial to ensure that the systems do \textbf{not} bias towards certain groups of people, such as preferring a certain gender for certain jobs.
Although both datasets used in this work already removed any sensitive information such as gender, we do not recommend directly deploying \framework{} for real-world applications without using debiasing techniques such as \citet{bolukbasi2016man,cheng2021fairfil,gaci-etal-2022-debiasing,guo-etal-2022-auto,schick-etal-2021-self}, and we do not condone the use of \framework{} for any morally unjust purposes.
To our knowledge, there is little work on investigating or mitigating biases in existing person-job fit systems, and we believe this is an important direction for future work.

\bibliography{anthology,custom}
\bibliographystyle{acl_natbib}

\clearpage
\appendix
\setcounter{table}{0}
\renewcommand{\thetable}{A\arabic{table}}
\setcounter{figure}{0}
\renewcommand{\thefigure}{A\arabic{figure}}
\section{More Details on Dataset and Preprocessing}
\label{sec:More Details on Dataset and Preprocessing}
\paragraph{Intellipro Dataset} The talent-job pairs come from the headhunting business in Intellipro Group Inc. The original resumes/job posts are parsed into text fields using techniques such as OCR. Some of the information is further corrected by humans. All sensitive information, such as names, contacts, college names, and company names, has been either removed or converted into numeric IDs. Example resume and job post are shown in \Cref{tbl:example_resume} and \Cref{tbl:example_job}, respectively.

\paragraph{AliYun Dataset} The 2019 Alibaba job-resume intelligent matching competition provided resume-job data that is already desensitized and parsed into a collection of text fields. There are 12 fields in a resume (\Cref{tbl:example_resume}) and 11 fields in a job post (\Cref{tbl:example_job}) used during training/validation/testing. Sensitive fields such as ``\chinese{居住城市}'' (living city) were already converted into numeric IDs. ``\chinese{工作经验}'' (work experience) was processed into a list of keywords. Overall, the average length of a resume or a job post in the AliYun dataset is much shorter than that of the Intellipro dataset (see \Cref{tbl:train_dset}). In our analysis, we also manually remapped the industries mentioned in the AliYun dataset into 20 categories such as ``Agriculture'', ``Manufacturing'', ``Financial Services'', etc., to be more comparable with the Intellipro dataset.
\section{More Details on Model Architecture}
\label{sec:more_details_on_model_architecture}

In this work, all data (resumes and job posts) are formatted as a collection of text fields. We simplify the model architecture from InEXIT \cite{InEXIT} to produce a single dense vector for a job post or a resume.
InEXIT first encodes each field independently using a pre-trained encoder, by encoding the field name (e.g., ``education'') and the value (e.g., ``Bachelor; major: Computer Science...) separately and then concatenating the two representations to obtain a representation for the entire field. Then, InEXIT models the ``internal interaction'' between fields from the same document using a self-attention layer. Next, InEXIT views resume-job matching as a non-linear interaction between the fields from a resume-job pair, and uses another self-attention to model the ``external interaction'' between all representations from both documents. Finally, InEXIT merges the representations obtained so far into a dense vector for a resume/job, concatenates the dense vectors to represent an resume-job pair, and finally uses an MLP layer to produce a matching score.

Compared to concatenating all text fields into a single string and using an encoder to directly produce an embedding, this approach of encoding each text field independently can effectively increase maximum context length (often 512). For example, we find fields such as ``Experiences'' and ``Projects'' in a resume from the Intellipro dataset often contain long texts. By encoding each field independently, we can include up to 512 tokens from \emph{each field}, compared to 512 tokens in total if the two fields are concatenated. We believe this is particularly suitable for modeling resume and job posts, as text fields (i.e., sections) from a resume/job post can be understood independently of other fields.

Since \framework{} models resume-job match using inner product (compatible with efficient retrieval frameworks such as FAISS \cite{FAISS}), we propose a few simplications to InEXIT's model architecture. First, since field names (e.g. ``education'', ``experiences'') are short, we directly concatenate them with the value to obtain a single string for each field (e.g., ``education: Bachelor in Computer Science, ...''). We then use a pre-trained encoder to directly obtain a representation for the entire field. Next, we follow InEXIT to use self-attention in a transformer layer to model the ``internal interaction'' between fields from the same document. After that, as we aim to model a resume and a job as dense vectors independent of each other, we remove the self-attention layer and the final MLP layer used to model a non-linear interaction between a resume-job pair. Instead, we use a linear layer to merge the representations for each text field, and output a dense vector for a resume or a job. This can then be used to perform inner product scoring, and can be combined with FAISS (see \Cref{subsec:Runtime}) to rank thousands of documents under miliseconds.

%
%
%
\begin{table}[!t]
  \centering
  \scalebox{0.8}{
    \begin{tabular}{l cc}
      \toprule
      Validation & \textbf{Intellipro Dataset} & \textbf{Aliyun Dataset}\\
      \midrule
      \# Jobs     & 109 & 299 \\
      \# Resumes  & 120 & 278 \\
      \# Labels   & 120 & 300 \\
      \bottomrule
    \end{tabular}
  }
\caption{Validation dataset statistics}
\label{tbl:val_dset}
\vspace{-10pt}
\end{table}
\begin{table*}[h]
  \centering
  \scalebox{0.8}{
    \begin{tabular}{p{0.6\linewidth} p{0.45\linewidth}}
      \toprule
      \multicolumn{1}{c}{\textbf{$R$ from Intellipro Dataset}} &
      \multicolumn{1}{c}{\textbf{$R$ from AliYun Dataset}} \\
      \midrule
      \underline{User ID}: xxxxx & 
      \underline{User ID}: xxxxx \\
      \underline{Languages}: ENGLISH; & \underline{\chinese{学历}}: \chinese{大专};\\
      \underline{Education}: start\_date: xxxx-xx-xx; & \underline{\chinese{年龄}}: 24;\\
      \phantom{\underline{Education}:} end\_date: xxxx-xx-xx; &\underline{\chinese{开始工作时间}}: 2018;\\
      \phantom{\underline{Education}:} college\_ranking: 20; &
      \underline{\chinese{居住城市}}: 551;\\
      \phantom{\underline{Education}:} major\_name: Computer Science; &
      \underline{\chinese{期望工作城市}}: 551,763,-;\\
      \phantom{\underline{Education}:} degree: BACHELOR; &
      \underline{\chinese{期望工作类型}}: \chinese{工程造价/预结算}\\
      \underline{Location}: city\_id: 115; &
      \underline{\chinese{期望工作行业}}: \chinese{房地产/建筑/工程}\\
      \phantom{\underline{Location}:} province\_id: 827; &
      \underline{\chinese{当前工作类型}}: \chinese{土木/建筑/装修/...}\\
      \phantom{\underline{Location}:} country\_id: 14; &
      \underline{\chinese{当前工作行业}}: \chinese{房地产/建筑/工程}\\
      \underline{Preferred Locations}: city\_id: 115; &
      \underline{\chinese{期望薪资}}: \chinese{xxxx-xxxx元/月}\\
      \phantom{\underline{Preferred Locations}:} province\_id: 827; &
      \underline{\chinese{当前薪资}}: \chinese{xxxx-xxxx元/月}\\
      \phantom{\underline{Preferred Locations}:} country\_id: 14; &
      \underline{\chinese{工作经验}}: \chinese{停车｜现场｜专家｜公园｜...}\\
      \underline{Industry}: SOFTWARE\_ENGINEERING; &
      \phantom{\chinese{工作经验}:} \textcolor{gray}{// other entries omitted}\\
      \underline{Skills}: azure; python; ... \textcolor{gray}{// other entries omitted} &
      \\
      \underline{Experiences}: title: Machine Learning Engineer; &
      \\
      \phantom{\underline{Experiences}:} start\_date: 2017-09; &
      \\
      \phantom{\underline{Experiences}:} end\_date: UNKNOWN; &
      \\
      \phantom{\underline{Experiences}:} company\_ranking: -1; &
      \\
      \phantom{\underline{Experiences}:} location: UNKNOWN; &
      \\
      \phantom{\underline{Experiences}:} description: Lead several MLOps projects... &
      \\
      \phantom{\underline{Experiences}:} title: Software Engineer; ... \textcolor{gray}{// other entries omitted} &
      \\
      \underline{Projects}: project\_name: xxxxx; &
      \\
      \phantom{\underline{Projects}:} title: Leader; &
      \\
      \phantom{\underline{Projects}:} start\_date: xxxx-xx-xx; &
      \\
      \phantom{\underline{Projects}:} end\_date: xxxx-xx-xx; &
      \\
      \phantom{\underline{Projects}:} description: Deploy template-based ... &
      \\
      \bottomrule
    \end{tabular}
  }
  \caption{Example resume from the Intellipro dataset and AliYun dataset. The Intellipro dataset contains resumes in both English and Chinese, while the AliYun dataset contains resumes only in Chinese. All documents are prepared as a collection of fields, displayed as: ``\underline{field name}: content''. Certain details are hidden for privacy concerns. \emph{User\_ID} is removed during training/validation/testing. Fields with multiple entries (e.g., \emph{Experiences} in the Intellipro dataset) are concatenated using newlines.}
  \label{tbl:example_resume}
\end{table*}
\begin{table*}[!ht]
  \centering
  \scalebox{0.8}{
    \begin{tabular}{p{0.65\linewidth} p{0.45\linewidth}}
      \toprule
      \multicolumn{1}{c}{\textbf{$J$ from Intellipro Dataset}} &
      \multicolumn{1}{c}{\textbf{$J$ from AliYun Dataset}} \\
      \midrule
      \underline{Job ID}: xxxxx &
      \underline{Job ID}: xxxxx\\
      \underline{Company Rank}: 12 &
      \underline{\chinese{工作名称}}: \chinese{工程预算}\\
      \underline{Company Description}: Energetic, exciting Silicon Valley startup. &
      \underline{\chinese{工作类型}}: \chinese{工程/造价/预结算}\\
      \underline{Job Title}: Deep Learning Specialist &
      \underline{\chinese{工作城市}}: \chinese{719}\\
      \underline{Job Location}: city\_id: 123; &
      \underline{\chinese{招聘人数}}: \chinese{3}\\
      \phantom{\underline{Job Location}:} province\_id: 335;&
      \underline{\chinese{薪资}}: \chinese{最低xxxx-最高xxxx元每月}\\
      \phantom{\underline{Job Location}:} country\_id: 56&
      \underline{\chinese{招聘开始时间}}: \chinese{2019xxxx}\\
       \underline{Job Position Type}: Full-time; &
      \underline{\chinese{招聘结束时间}}: \chinese{2019xxxx}\\
      \underline{Job Description/Responsibilities}: Use computer vision, computational geometry, and ... \textcolor{gray}{// other details omitted} &
      \underline{\chinese{工作描述}}: \chinese{工程预算员岗位职责：1.能够独立完成...}\textcolor{gray}{// other details omitted}\\
      \underline{Required Qualifications/Skills}: Strong programming experience in Python, C++, or Java; PhD in Computer Science, Electrical Engineering, ... \textcolor{gray}{// other details omitted}& \underline{\chinese{最低学历}}: \chinese{大专}\\
      \underline{Preferred Qualifications/Skills}: UNKNOWN & \underline{\chinese{是否要求出差}}: \chinese{0}\\
      & \underline{\chinese{工作年限}}: \chinese{五年到十年}\\
      \bottomrule
    \end{tabular}
  }
  \caption{Example job posts from the Intellipro dataset and AliYun dataset. The Intellipro dataset contains job posts in both English and Chinese, while the AliYun dataset contains job posts only in Chinese. All documents are prepared as a collection of fields, displayed as: ``\underline{field name}: content''. Certain details are omitted. \emph{Job\_ID} is removed during training/validation/testing.}
  \label{tbl:example_job}
\end{table*}
\section{More Details on Baselines}
\label{sec:more_details_on_baselines}

\paragraph{XGBoost} We use ``XGBoost-classifier'' \cite{XGBoost} for classification based metrics, and ``XGBoost-ranker'' for ranking based metrics in \Cref{tbl:main_exp_bert} and \Cref{tbl:main_exp_e5}. Similar to other classification-targeted methods such as \emph{MV-CoN} and \emph{InEXIT}, we use $\mathcal{D}$ without ``contrastive learning''. Hyperparameters are tuned using grid search, and classification thresholds are found using the validation set.

\paragraph{RawEmbed} We first concatenate all fields in a resume/job post into a single string, and use pre-trained encoders such as BERT \cite{devlin2019bert}, E5 \cite{e5}, xlm-roberta \cite{xlm-roberta}, and OpenAI text-ada-002 \cite{text-ada} to produce a dense embedding. We use inner product to produce a score for ranking tasks, and use cosine similarity with a threshold found using the validation set for classification tasks.

\paragraph{MV-CoN} We follow the official implementations from \citet{mvcon}, but replace the fixed embedding layer with the architecture shown in \Cref{subsec:Model Architecture} and \Cref{fig:model_archi}, since our test set considers \emph{unseen} resumes and job posts. We use AdamW optimizer \cite{adamw} with a learning rate of 5e-6, a linear warm-up schedule for the first 10\% of the training steps, and a weight decay of 1e-2 for both datasets. We use a batch size of 4 with a gradient accumulation of 4 when a small encoder (e.g., BERT-base) is used, and use DeepSpeed Zero 2 \cite{deepspeed} with BF16 mixed precision training when a large encoder (e.g., E5-large) is used.

\paragraph{InEXIT} We follow the official implementation from \citet{InEXIT} to model both the ``internal'' and ``external'' interaction between a resume-job pair. We use AdamW optimizer \cite{adamw} with a learning rate of 5e-6, a linear warm-up schedule for the first 10\% of the training steps, and a weight decay of 1e-2 for both datasets. We use a batch size of 8 with a gradient accumulation of 2 when a small encoder is used, and a batch size of 4 with a gradient accumulation of 4 when a large encoder is used.\footnote{
  In our experiment, we find that {InEXIT} \cite{InEXIT} performs slightly worse than MV-CoN \cite{mvcon} on the AliYun dataset (see \Cref{tbl:main_exp_bert}), while \citet{InEXIT} reports the contrary. We believe this is because {InEXIT} considers a test setting where part of the resumes/job posts can be \emph{seen} in training, since training/validation/testing pairs are simply randomly sampled. In contrast, in our experiment, we consider  test and validation set with only resumes/job posts \emph{not seen} during training.
}

\paragraph{DPGNN} We follow the official implementation from \citet{DPGNN}, but remove the fixed-size embedding layer in the graph neural network for encoding a resume or a job, since our test set considers \emph{unseen} resumes and job posts. We replace the embedding layer with a pre-trained encoder (e.g., BERT), and keep other aspects the same, such as modeling both the ``active'' and ``passive'' representation of a resume or a job post. We also removed the GraphCNN module as we do not have ``interaction records'' (e.g., \emph{recruiters reaching out} to job seekers) used to train this module, and the total number of labels in our resume-job datasets is also small. Finally, we modified the proposed BPR loss \cite{DPGNN} by first normalizing all embedding vectors, since we found training DPGNN with the original BPR loss results in high numerical instability. We use AdamW optimizer \cite{adamw} with a learning rate of 1e-5, a linear warm-up schedule for the first 5\% of the training steps, and a weight decay of 1e-2 for both datasets. We use a batch size of 8 with a gradient accumulation of 2 when using a small encoder, and a batch size of 4 with a gradient accumulation of 4 when using a large encoder.

\paragraph{BM25} Since resumes in the Intellipro dataset can be long, we use BM25L \cite{bm25l,bm25-all} for ranking tasks. We use the implementation from \citet{rank_bm25} with the default hyperparameters.

In general, all neural-network-related code is implemented using PyTorch Lightning \cite{Falcon_PyTorch_Lightning_2019}, and all training is performed on a single A100 80GB GPU. We train all models for 10 epochs and save the best checkpoint based on validation loss for testing.
On average, it takes about 1 hour and 4 hours to train \emph{MV-CoN}, \emph{InEXIT}, \emph{DPGNN} using a small encoder on the Intellipro dataset and the AliYun dataset, respectively. When using a large encoder (e.g., E5-large), it takes about 5-8 hours and 19-24 hours to train on the Intellipro dataset and the AliYun dataset, respectively.



\section{\framework{} Training Hyperparameters}
\label{sec:Training Hyperparameters}

In general, \framework{} first performs data augmentation using both ChatGPT and EDA (see \Cref{subsec:Data Augmentation} and \Cref{sec:More Details on Data Augmentation} for more details), and then trains the model architecture shown in \Cref{fig:model_archi} using contrastive learning (see \Cref{subsec:Contrastive Learning}). Similar to baseline methods (see \Cref{sec:more_details_on_baselines}), we use the AdamW optimizer \cite{adamw}, a linear warm-up schedule for the first 5\% of the training steps, and a weight decay of 1e-2 for both datasets. We use a batch size of $B=8, B_{\mathrm{hard}}=8$ with a gradient accumulation of 2 when using a small encoder for both datasets. When using a large encoder (e.g., E5-large) on the Intellipro dataset, we keep the same batch size of $B=8$, but with $B_{\mathrm{hard}}=4$ and DeepSpeed Zero 2 \cite{deepspeed} with BF16 mixed precision training due to GPU memory constraints. On the AliYun dataset, we simply use $B=8, B_{\mathrm{hard}}=8$ without DeepSpeed as input sequences are much shorter compared to those from the Intellipro dataset.

We train \framework{} models for 10 epochs and save the best checkpoint based on validation loss for testing. On average, \framework{} takes about 1.5 hours and 4.5 hours to train when using a small encoder on the Intellipro dataset and the AliYun dataset, respectively. When using a large encoder (e.g., E5-large), \framework{} takes about 3 hours and 9 hours to train on the Intellipro dataset and the AliYun dataset, respectively.
\begin{figure*}[ht!]
  \centering
  \subfigure[\framework{}]{%
      \label{fig:viz_confit_appendix}%
  \includegraphics[scale=0.26]{./images/confit_raw_0911_bert_resume.pdf}}%
  \subfigure[E5-small]{%
      \label{fig:viz_e5-small_appendix}%
  \includegraphics[scale=0.26]{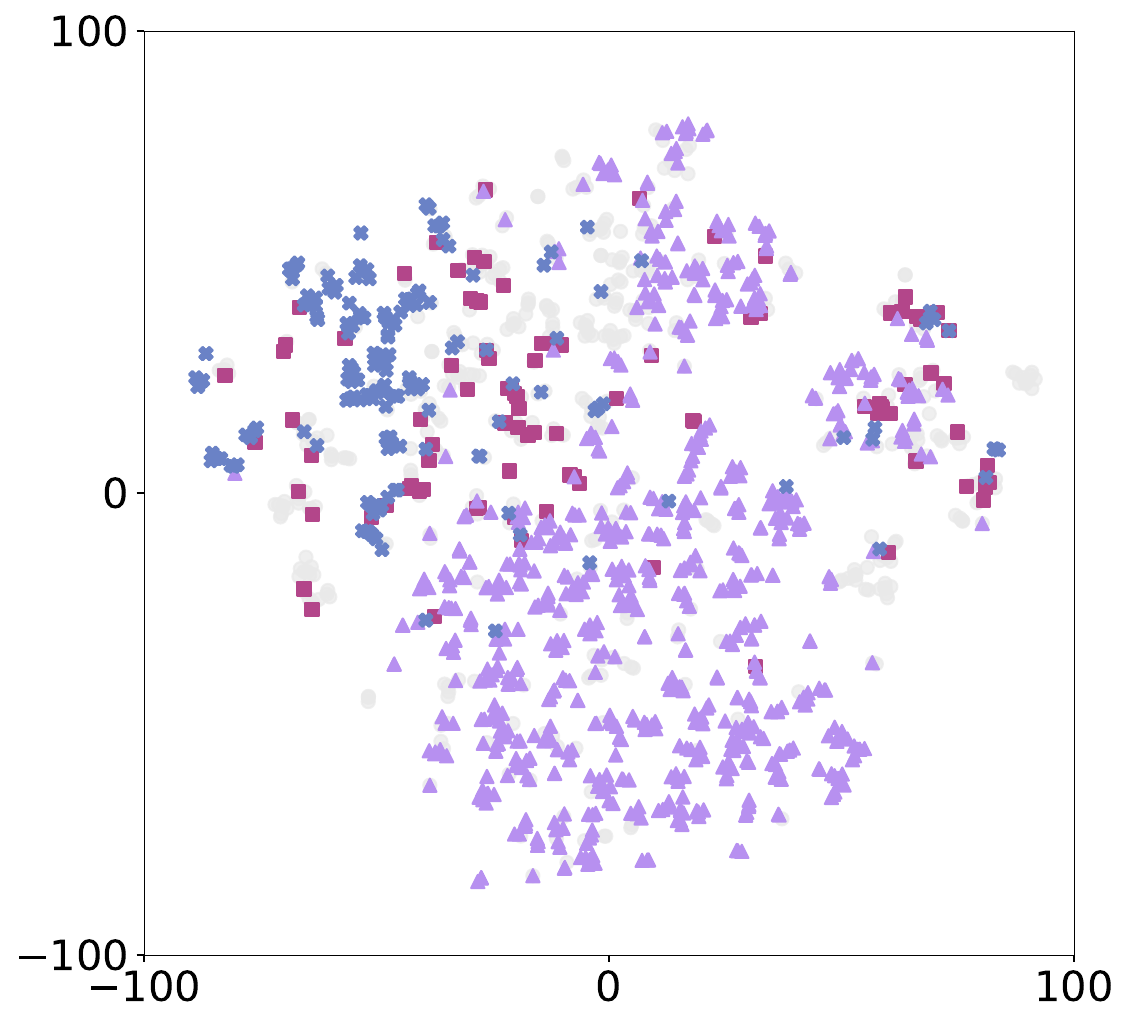}}%
  \subfigure[text-ada-002]{%
      \label{fig:viz_openai_appendix}%
  \includegraphics[scale=0.26]{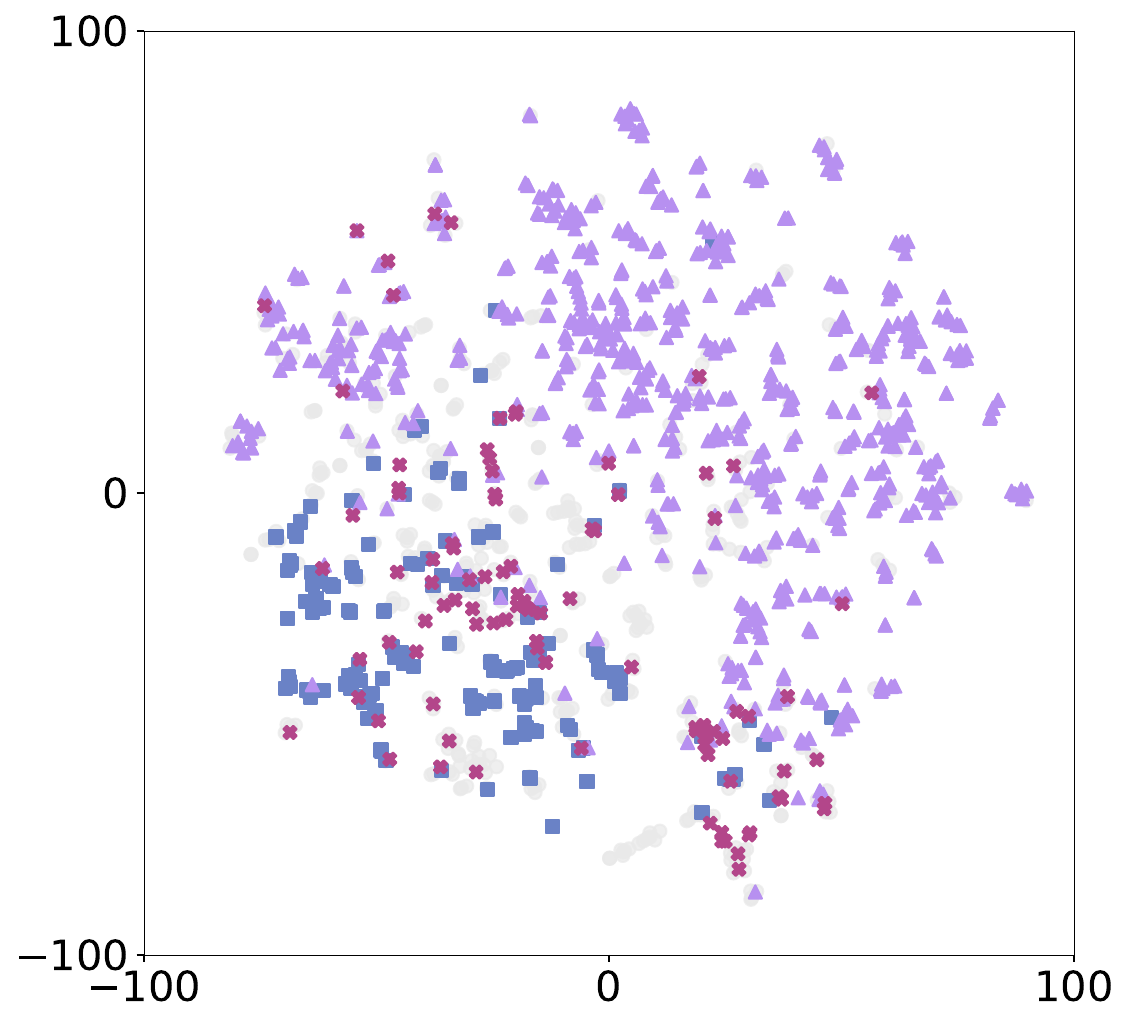}}%
  \\
  \subfigure[BERT-base]{%
      \label{fig:viz_bert-multilingual_appendix}%
  \includegraphics[scale=0.26]{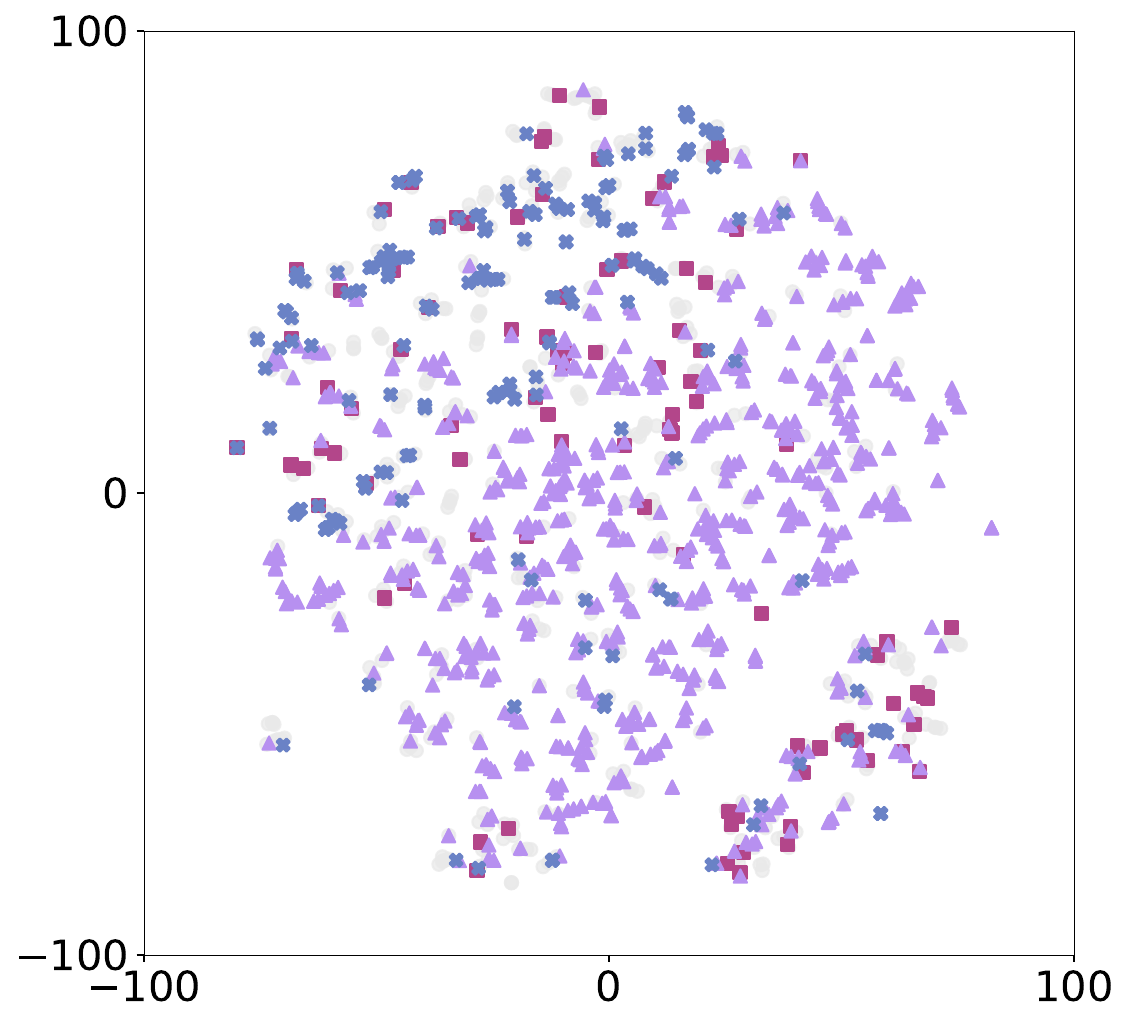}}%
  \subfigure[MV-CoN]{%
      \label{fig:viz_mvcon_appendix}%
  \includegraphics[scale=0.26]{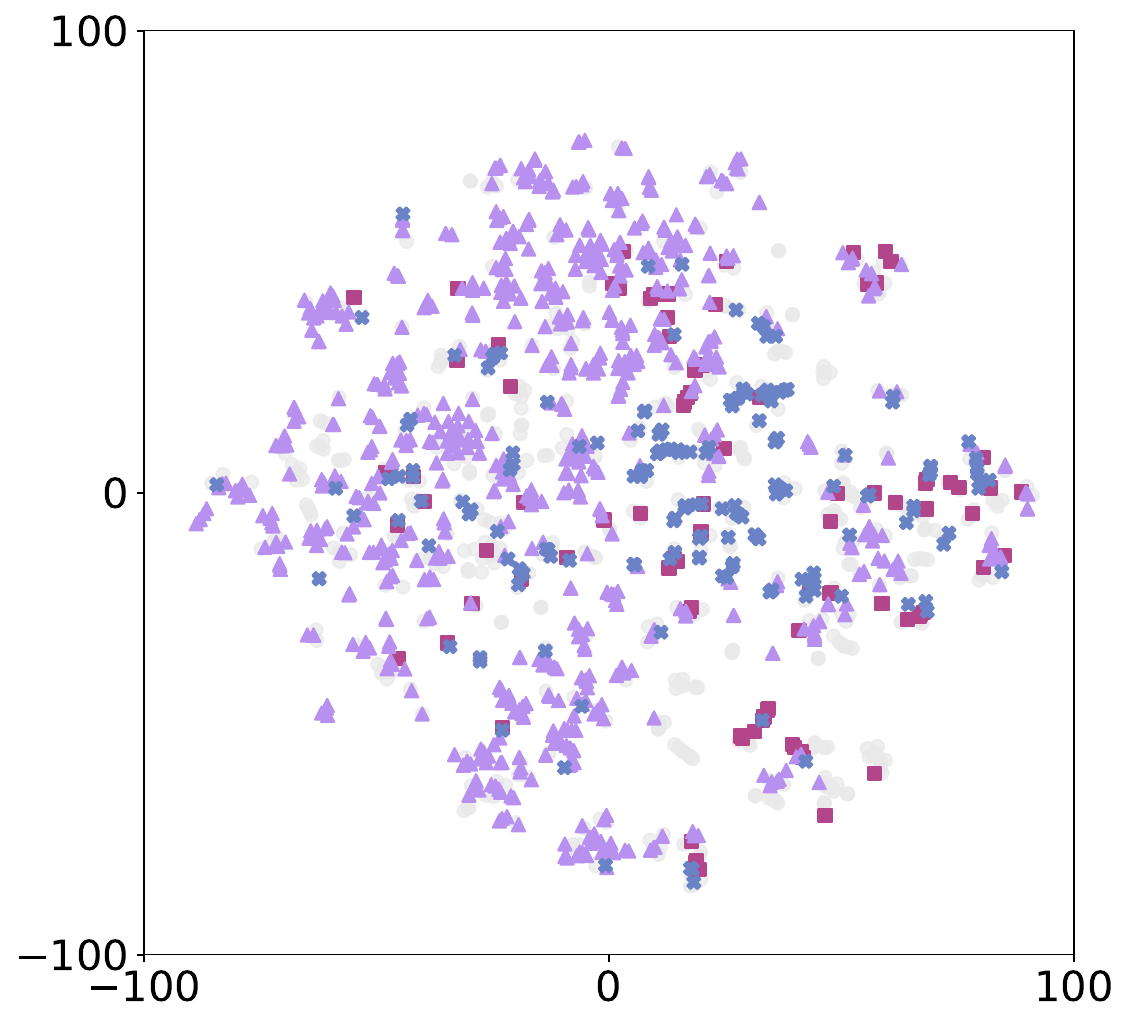}}%
  \subfigure[DPGNN]{%
      \label{fig:viz_dpgnn_appendix}%
  \includegraphics[scale=0.26]{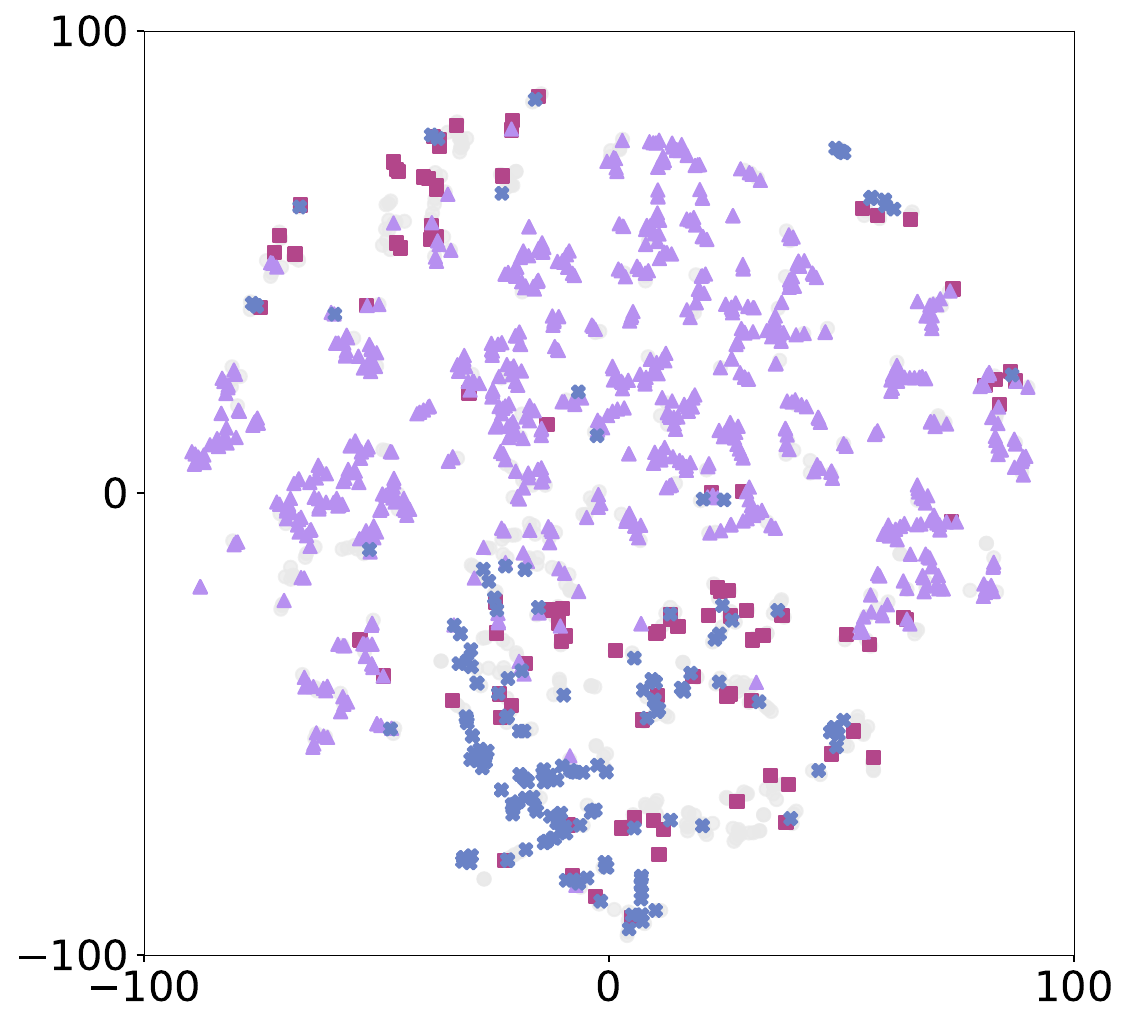}
  }%
  \caption{Resume embeddings produced by various methods in \Cref{tbl:main_exp_bert} with BERT-base-multilingual-cased as backbone encoder. Colors assigned using each resume's desired industry. Top-3 most frequent industries are color-coded for easier viewing. \emph{BERT-base} refers to raw embedding produced by BERT-base-multilingual-cased.}
  \label{fig:viz_all}
\end{figure*}
\begin{table*}[!t]
  \centering
  \scalebox{0.67}{
    \begin{tabular}{ll ccc cccc ccc cccc}
      \toprule
      & & \multicolumn{7}{c}{\textbf{Intellipro Dataset}} & \multicolumn{7}{c}{\textbf{AliYun Dataset}} \\
      & & \multicolumn{2}{c}{Rank Resume} & \multicolumn{2}{c}{Rank Job} & \multicolumn{3}{c}{Classification} 
      & \multicolumn{2}{c}{Rank Resume} & \multicolumn{2}{c}{Rank Job} & \multicolumn{3}{c}{Classification} \\
      \cmidrule(lr){3-4} \cmidrule(lr){5-6} \cmidrule(lr){7-9} 
      \cmidrule(lr){10-11} \cmidrule(lr){12-13} \cmidrule(lr){14-16}
      \textbf{Method} & \textbf{Encoder} 
      & MAP & nDCG & MAP & nDCG & F1 & Prc+ & Rcl+ 
      & MAP & nDCG & MAP & nDCG & F1 & Prc+ & Rcl+ \\
      \midrule
      MV-CoN
      & BERT-base
      & 10.81 & 10.00 & 3.34 & 2.17
      & \second{\textbf{58.00}} & \second{\textbf{50.00}} & 33.33 
      & 5.41  & 5.15  & 13.44 & 12.67
      & \textbf{74.25} & \textbf{72.22} & \second{\textbf{68.32}}\\
      InEXIT
      & BERT-base
      & 12.27 & 12.98 & 4.11 & 3.46
      & 55.55 & 44.74 & 35.42 
      & 5.25  & 4.98  & 13.02 & 12.30
      & \second{\textbf{71.75}} & \second{\textbf{66.67}} & \textbf{72.18} \\
      DPGNN
      & BERT-base
      & \second{\textbf{19.64}} & \second{\textbf{21.95}} & \textbf{17.86} & \textbf{19.60}
      & \textbf{61.16} & \textbf{52.38} & \second{\textbf{45.83}} 
      & \second{\textbf{19.96}} & \second{\textbf{24.64}} & \second{\textbf{27.23}} & \second{\textbf{30.07}}
      & 50.31 & 45.24 & 57.14 \\
      \cmidrule(lr){2-16}
      Ours+XGBoost
      & TF-IDF &
    \textbf{24.04} & \textbf{27.29} & \second{\textbf{15.60}} & \second{\textbf{17.23}} &
    43.60 & 37.66 & \textbf{60.42} &
    \textbf{24.19} & \textbf{28.95} & \textbf{30.29} & \textbf{33.66} &
    52.31 & 47.51 & 64.67 \\
      \bottomrule
    \end{tabular}
  }
  \caption{\framework{} without neural networks (denoted as \emph{Ours+XGBoost}) is competitive against many prior person-job fit methods with BERT-base as a backbone encoder. \emph{F1} is weighted F1 score, \emph{nDCG} is nDCG@10, \emph{Prc+} and \emph{Rcl+} are precision and recall for positive classes. Results for non-deterministic methods are averaged over 3 runs. Best result is shown in \textbf{bold}, and runner-up is in \second{\textbf{gray}}.}
  \label{tbl:main_exp_xgboost}
  \vspace{-5pt}
\end{table*}
\section{More Details on Data Augmentation}
\label{sec:More Details on Data Augmentation}

In \Cref{subsec:Data Augmentation}, we discussed how \framework{} can increase the number of resume-job labels by first creating augmented resumes $\hat{R}_i$ and jobs $\hat{J}_i$ that carry semantically similar information as $R_i$ and $J_i$, and then replicating the labels from $R_i$ and $J_i$ to $\hat{R}_i$ and $\hat{J}_i$. Since much information in a resume or a job post contains formal names such as ``Job Title'', we \emph{only paraphrase certain sections}. For resumes in the Intellipro dataset, we paraphrase the ``description'' subsection in the ``Experiences'' section and the ``description'' subsection in the ``Projects'' section (see \Cref{tbl:example_resume}). For job posts in the Intellipro dataset, we paraphrase the ``Company Description'' section, the ``Job Description/Responsibilities'' section, the ``Required Qualifications/Skills'', and the ``Preferred Qualifications/Skills'' section (see \Cref{tbl:example_job}). For the AliYun dataset, we paraphrase the ``\chinese{工作经验}'' (work experience) section for resumes, and the ``\chinese{工作描述}'' (job description) section for job posts.

\framework{} performs data augmentation using both ChatGPT and EDA for 500 resumes and 500 jobs for each dataset. With only 1000 augmented documents on each dataset, we increased the number of resume-job labels by 5330 and 9706 for the Intellipro dataset and the AliYun dataset, respectively.

\section{More Details on Runtime Comparison}
\label{sec:More Details on Runtime Comparison}

In \Cref{subsec:Runtime}, we compared the runtime of various neural-based methods from \Cref{tbl:main_exp_bert}. We categorize neural-based methods into two types when doing inference: Maximum Inner Product Search (\emph{MIPS}) methods and Non-linear (\emph{Non-linear}) methods. MIPS methods compute a matching score between two dense vectors using inner product, and can be efficiently implemented using FAISS \cite{FAISS} to scale to billions of documents. MIPS-based approach includes \emph{RawEmbed} and \framework{}.
Non-linear methods produce a matching score by modeling non-linear interactions between a resume and a job's (intermediate) representations. For example, \emph{InEXIT} first concatenates the intermediate representations of a resume and a job, and then passes them into a self-attention layer and an MLP layer for scoring. Non-linear methods include \emph{MV-CoN}, \emph{InEXIT}, and \emph{DPGNN}.

All experiments are performed using the test set from the AliYun dataset on a single A100 80GB GPU. For MIPS-based methods, we precompute all the relevant embeddings (excluded from runtime calculation), and record the average runtime for FAISS to retrieve the top 10 job posts from a pool of 100, 1000, and 10000 job posts when given a resume embedding. For non-linear methods, we record the average runtime to perform all the needed forward passes for each of the 100, 1000, and 10000 resume-job \emph{pairs}. However, we do note that the runtime for non-linear methods \emph{can be further optimized} by precomputing certain intermediate representations before passing them into their respective non-linear scoring layers. We did not perform this optimization because 1) this is highly architecture- and method-dependent, and 2) it still does not scale well when the number of job posts is large, or when there are multiple resumes to query.

\section{More Details on Ablation Studies}
\label{sec:More Details on Ablation Studies}

Our ablation studies in \Cref{subsec:Ablation Studies} also experimented with removing neural networks completely, to decouple our methodology from any particular choice of neural networks. To achieve this, we first mimic the batches used during contrastive training in \framework{} and construct a dataset $\mathcal{D}_{\mathrm{con}}$ which contains a positive resume-job pair $\langle R^{+}_i, J^{+}_i \rangle$ along with $l$ negative resumes and $l$ negative job posts (see \Cref{subsec:Contrastive Learning}). Then, we treat all negative resumes and job posts that have a label of $y=0$ when paired with $J^{+}_i$ and $R^{+}_i$, respectively. Finally, we encode all resumes and job posts using TF-IDF, and train an XGBoost ranker using $\mathcal{D}_{\mathrm{con}}$. To be comparable with \framework{} which uses $B=8, B_{\mathrm{hard}}=8$, we use $l=16$ for each positive resume-job pair, with 14 random negatives and 2 hard negatives.

We denote this approach as \emph{Ours+XGboost}, and compare its performance against other person-job fit systems in \Cref{tbl:main_exp_xgboost}. We find our approach is still competitive against these methods that use a BERT-base \cite{devlin2019bert} encoder. This suggests that the contrastive learning and data augmentation procedure from \framework{} is effective for the person-job fit task.


\section{More Details on Qualitative Analysis}
\label{sec:More Qualitative Analysis}

\Cref{fig:viz_all} presents the resume embeddings produced by various methods in \Cref{tbl:main_exp_bert} with BERT-base-multilingual-cased as the backbone encoder (with the exception of OpenAI text-ada-002, which is from \Cref{tbl:main_exp_e5}). Since methods such as \emph{MV-CoN}, \emph{InEXIT}, and \emph{DPGNN} does not explicitly learn a resume or a job embedding, we extract the representations from the last layer before their resume-job pair scoring layers (e.g., the final MLP layer in \emph{MV-CoN}, or the self-attention layers in \emph{InEXIT}).

In general, we find embeddings produced by \emph{MV-CoN}, \emph{DPGNN}, and \emph{BERT-base} tend to scatter ``Software Engineering''-related resumes across the entire embedding space, while embeddings produced by \framework{}, \emph{E5-small}, and \emph{text-ada-002} has a clearer separation between ``Software Engineering'' and other industries such as ``Human Resource''. In \Cref{tbl:main_exp_bert}, we similarly find the ranking performances of \framework{}, \emph{E5-small}, and \emph{text-ada-002} are better than \emph{MV-CoN}, \emph{DPGNN}, and \emph{BERT-base} on the Intellipro dataset. Therefore, we believe \Cref{fig:viz_all} qualitatively shows that having a high-quality embedding space is beneficial for modeling person-job fit.

\end{document}